\documentclass[draftcls, onecolumn, journal]{IEEEtran}
\pdfoutput=1

\ifCLASSINFOpdf
   \usepackage[pdftex]{graphicx}
   \graphicspath{{./pdfFig/}{../jpegFig/}}
   \DeclareGraphicsExtensions{.pdf,.jpeg,.png}
\else
\fi
\usepackage[cmex10]{amsmath}
\usepackage[tight,footnotesize]{subfigure}

\usepackage{color}
\usepackage{enumerate}

\usepackage{color}
\usepackage{amsmath}
\usepackage{amssymb}
\usepackage{graphicx}
\usepackage{comment,xspace}
\usepackage{algorithm,algorithmic} 
\usepackage{fancybox}

\newcommand{\real}{{\mathbb{R}}}
\newcommand{\reals}{\real}

\newtheorem{theorem}{Theorem}[section]

\newtheorem{lemma}[theorem]{Lemma}
\newtheorem{remark}[theorem]{Remark}
\newtheorem{example}[theorem]{Example}
\newtheorem{definition}[theorem]{Definition}
\newtheorem{corollary}[theorem]{Corollary}

\newcommand{\expectation}[1]{\mbox{$\mathbb{E}\left[#1\right]$}}

\def\A{\mathcal{A}}
\def\C{\mathcal{C}}
\def\F{\mathcal{F}}

\def\A{\mathcal{A}}

\begin{document}
%
\title{Equitable Partitioning Policies\\ for Mobile Robotic Networks}
%
%
%
\author{Marco~Pavone,~Alessandro~Arsie,~Emilio~Frazzoli,~Francesco~Bullo
\thanks{Marco Pavone and Emilio Frazzoli are with the Laboratory for Information and Decision Systems, Department of Aeronautics and
Astronautics, Massachusetts Institute of Technology, Cambridge, {\tt\small \{pavone, frazzoli\}@mit.edu}.}
\thanks{Alessandro Arsie is with the Department of Mathematics, Pennsylvania State University, {\tt\small arsie@math.psu.edu}.}%
\thanks{Francesco Bullo is with the Center for Control Engineering and Computation, University of California at Santa Barbara, {\tt\small bullo@engineering.ucsb.edu}.}}

%
%

\markboth{IEEE TRANSACTIONS ON AUTOMATIC CONTROL, SUBMITTED FOR PUBLICATION AS A FULL PAPER.}%
{}

%



\maketitle

\begin{abstract}
The most widely applied strategy for workload sharing is to equalize the workload assigned to each resource. In mobile multi-agent systems, this principle directly leads to equitable partitioning policies in which (i) the workspace is divided into subregions of equal measure, (ii) there is a bijective correspondence between agents and subregions, and (iii) each agent is responsible for service requests originating within its own subregion. In this paper, we design provably correct, spatially-distributed and adaptive policies that allow a team of agents to achieve a convex and equitable partition of a convex workspace, where each subregion has the same measure. We also consider the issue of achieving  convex and equitable partitions where subregions have shapes similar to those of regular polygons. Our approach is related to the classic Lloyd algorithm, and exploits the unique features of power diagrams. We discuss possible applications to routing of vehicles in stochastic and dynamic environments. Simulation results are presented and discussed. 
\end{abstract}

\section{Introduction}
In the near future, large groups of autonomous agents will be used to perform complex tasks including transportation and distribution, logistics, surveillance, search and rescue operations, humanitarian demining, environmental monitoring, and planetary
exploration. The potential advantages of multi-agent systems are, in fact, numerous. For instance, the intrinsic parallelism of a multi-agent system provides robustness to failures of single agents, and in many cases can guarantee better time efficiency. Moreover, it is possible to reduce the total implementation and operation cost, increase reactivity and system reliability, and add flexibility and modularity to monolithic approaches.

In essence, agents can be interpreted as \emph{resources} to be \emph{shared} among \emph{customers}. In surveillance and exploration missions, customers are points of interests to be visited; in transportation and distribution applications, customers are people demanding some service (e.g., utility repair) or goods; in logistics tasks, customers could be troops in the battlefield. Finally, consider a possible architecture for
networks of  autonomous agents performing distributed sensing: a set of $n$ cheap sensing devices (sensing nodes), distributed in the environment, provides sensor measurements, while $m$ sophisticated agents (cluster heads) collect information from the sensing nodes and transmit it (possibly after some computation) to the outside world. In this case, the sensing nodes represent customers, while the agents, acting as cluster heads, represent resources to be allocated.

The most widely applied strategy for workload sharing among resources is to equalize the
total workload assigned to each resource. While, in principle, several
strategies are able to guarantee workload-balancing in multi-agent systems,
\emph{equitable partitioning policies} are predominant
\cite{Bertsimas.vanRyzin.Demand:93, Baron.Berman.ea:07, Liu.Liu.ea:03,
  Carlsson.Ge.ea:07}.  A partitioning policy is an algorithm that, as a
function of the number $m$ of agents and, possibly, of their position and
other information, partitions a bounded workspace $A \subset \reals^d$ into $m$ openly disjoint
regions $A_i$, for $i\in\{1,\dots,m\}$.  (Voronoi diagrams are an example
of a partitioning policy.) In the resource allocation problem, each agent $i$
is assigned to subregion $A_i$, and each customer in $A_i$ receives service
by the agent assigned to $A_i$.  Accordingly, if we model the
\emph{workload} for subregion $S \subseteq A$ as $\lambda_{S} \doteq \int_{
  S} \lambda (x) \, dx$, where $\lambda(x)$ is a measure over $A$, then the
workload for agent $i$ is $\lambda_{A_i}$.  Given this preface,
load-balancing calls for equalizing the workload $\lambda_{A_i}$ in the $m$
subregions or, in equivalent words, to compute an \emph{equitable}
partition of the workspace $A$ (i.e., a partition where $\lambda_{A_i} =
\lambda_A/m$, for all $i$).



Equitable partitioning policies are predominant for three main reasons: (i) efficiency, (ii) ease of design and (iii) ease of analysis. Equitable partitioning policies are, therefore, ubiquitous in multi-agent system applications. To date, nevertheless, to the best of our knowledge, all equitable partitioning policies inherently assume a \emph{centralized} computation of the workspace partition. This fact is in sharp contrast with the desire of a fully distributed architecture for a multi-agent system. The lack of a fully distributed architecture limits the applicability of equitable partitioning policies to limited-size multi-agent systems operating in a known static environment.

The contribution of this paper is three-fold. First, we design provably correct, spatially-distributed, and adaptive policies that allow a team of agents to achieve a convex and equitable partition of a convex workspace. Our approach is related to the classic Lloyd
algorithm from vector quantization theory~\cite{Lloyd:82, FB-JC-SM:08}, and exploits the unique
features of power diagrams, a generalization of Voronoi diagrams (see
\cite{Kwok.Martinez:07} for another interesting application of power
diagrams in mobile sensor networks). Second, we provide extensions of our algorithms to take into account \emph{secondary} objectives, as for example, \emph{control on the shapes of the subregions}. Our motivation, here, is that equitable partitions in which subregions are thin slices are, in most applications, impractical: in the case of dynamic vehicle routing, for example, a thin slice partition would directly lead to an increase in fuel consumption. Third, we discuss some applications of our algorithms;
we focus, in particular, on the Dynamic Traveling Repairman Problem (DTRP)~\cite{Bertsimas.vanRyzin.Demand:93}, where equitable partitioning policies are indeed optimal under some assumptions.

Finally, we mention that our algorithms, although motivated in the context of multi-agent systems, are a novel contribution to the field of computational geometry. In particular we address, using a dynamical system framework, the well-studied equitable convex partition problem (see \cite{Carlsson.Armbruster.ea:08} and references therein); moreover, our results provide new insights in the geometry of Voronoi diagrams and power diagrams.

The paper is organized as follows. In Section II we provide 
the necessary tools from calculus, degree theory and geometry. Section III contains the problem formulation, while in Section IV we present preliminary algorithms for equitable partitioning based on leader-election, and we discuss their limitations. Section V is the core of the paper: we first prove some existence results for power diagrams, and then we design provably correct, spatially-distributed, and adaptive equitable partitioning policies that do not require any leader election. In Section VI we extend the algorithms developed in Section V to take into account \emph{secondary} objectives. Section VII contains simulations results. Finally, in Section VIII, we provide an application of our algorithms to the DTRP problem, and we draw our conclusions.

\section{Background}
\label{sec:Background}

In this section, we introduce some notation and  briefly review some  concepts from calculus, degree theory and geometry, on which we will rely extensively later in the paper.

\subsection{Notation}
Let $\|\cdot \|$ denote the Euclidean norm. Let $A$ be a compact, convex subset of $\reals^d$. We denote the boundary of $A$ as $\partial A$ and the Lebesgue measure of $A$ as $|A|$. We define the diameter of $A$ as:  $\mathrm{diameter}(A) \doteq \sup \{ ||p-q ||\, |\, p,q \in A\}$. The distance from a
point $x$ to a set $M$ is defined as $\textrm{dist}(x,M) \doteq \inf_{p \in M}\|x - p \|$. We define $I_m\doteq  \{1,2,\cdots,m \}$. Let $G = (g_1,\cdots, g_m ) \subset A^m$ denote the location of $m$ points. A \emph{partition} (or tessellation) of $A$ is a collection of $m$ closed subsets $\mathcal A = \{A_1,\cdots,A_m \}$ with disjoint interiors whose union is $A$. A partition $\mathcal{A}=\{A_1, \cdots, A_m\}$ is \emph{convex} if each  $A_i$, $i \in I_m$, is convex.

Given a vector space $\mathbb{V}$, let $\mathbb{F}(\mathbb{V})$ be the collection of finite subsets of $\mathbb{V}$. Accordingly, $\mathbb{F}(\reals^d)$ is the collection of finite point sets in $\reals^d$. Let $\mathbb{G}(\mathbb{R}^d)$ be the set of undirected graphs whose vertex set is an 
element of $\mathbb{F}(\reals^d)$ (we assume the reader is familiar with the 
standard notions of graph theory as defined, for instance, in \cite[Chapter 1]{Diestel:00}).

Finally, we define the saturation function $\text{sat}_{a,b}(x)$, with $a<b$, as:
\begin{equation}
\text{sat}_{a,b}(x) = \left\{ \begin{array}{ll}
1 & \textrm{if $x>b$}\\
(x-a)/(b-a) & \textrm{if $a\leq x \leq b$}\\
0 & \textrm{otherwise}
\end{array} \right.
\end{equation}

\subsection{Variation of an Integral Function due to a Domain Change.}
The following result is related to classic divergence theorems \cite{Chavel:84}. Let $\Omega = \Omega(y) \subset A$ be a region that depends smoothly on a real parameter $y \in \reals$ and that has a well-defined boundary $\partial \Omega(y)$ for all $y$. Let $h$ be a density function over $A$. Then
\begin{equation}
\label{eq:divergence}
\frac{d}{dy}\int_{\Omega(y)} h(x) \, dx = \int_{\partial \Omega(y)} \Bigl ( \frac{dx}{dy} \cdot n(x) \Bigr ) \, h(x) \,dx,
\end{equation}
where $v\cdot w$ denotes the scalar product between vectors $v$ and $w$, where $n(x)$ is the unit outward normal to $\partial \Omega(y)$, and where $dx/dy$ denotes the derivative of the boundary points with respect to $y$.

\subsection{A Basic Result in Degree Theory}

In this section, we state some results in degree theory that will be useful in the remainder of the paper. For a thoroughly introduction to the theory of degree we refer the reader to  \cite{Hatcher:01}. 

Let us just recall the simplest definition of degree of a map $f$. Let $f: X\rightarrow Y$ be a smooth map between connected compact manifolds $X$ and $Y$ of the same dimension, and let $p\in Y$ a regular value for $f$ (regular values abound due to Sard's lemma). Since $X$ is compact, $f^{-1}(p)=\{x_1,\dots, x_n\}$ is a finite set of points and since $p$ is a regular value, it means that $f_{U_i}: U_i \rightarrow f(U_i)$ is a local diffeomorphism, where $U_i$ is a suitable open neighborhood of $x_i$. Diffeomorphisms can be either orientation preserving or orientation reversing. Let $d^+$ be the number of points $x_i$ in $f^{-1}(p)$ at which $f$ is orientation preserving (i.e. $\mathrm{det}(\mathrm{Jac}(f))>0$, where $\mathrm{Jac}(f)$ is the Jacobian matrix of $f$) and $d^{-}$ be the number of points in $f^{-1}(p)$ at  which f is orientation reversing (i.e. $\mathrm{det}(\mathrm{Jac}(f))<0$). Since $X$ is connected, it can be proved that the number $d^+ -d^-$ is independent on the choice of $p\in Y$ and one defines $\mathrm{deg}(f):=d^+ - d^-$. The degree can be also defined for a {\em continuous} map $f: X \rightarrow Y$ among connected oriented topological manifolds of the same dimensions, this time using homology groups or the local homology groups at each point in $f^{-1}(p)$ whenever the set $f^{-1}(p)$ is finite. For more details see \cite{Hatcher:01}.



The following result will be fundamental to prove some existence theorems and it is a direct consequence of the theory of degree of continuous maps among spheres,
\begin{theorem}\label{eq:homotopy}
Let $f: B^n \rightarrow B^n$ be a continuous map from a closed $n$-ball to itself. Call $S^{n-1}$ the boundary of $B^n$, namely the $(n-1)$-sphere and assume that $f_{S^n}: S^n \rightarrow S^n$ is a map with $\mathrm{deg}(f)\neq 0$. Then $f$ is {\em onto} $B^n$. 
\end{theorem}
\begin{proof}
Since $f$ as a map from $S^{n-1}$ to $S^{n-1}$ is different from zero, then the map $f_{S^{n-1}}$ is onto the sphere. If $f$ is not onto $B^n$, then it is homotopic to a map  $B^n \rightarrow  S^{n-1}$, and then $f_{S^{n-1}}: S^{n-1}   \rightarrow S^{n-1}$ is homotopic to the trivial map (since it extends to the ball). Therefore $f_{S^{n-1}}: S^{n-1}   \rightarrow S^{n-1}$ has zero degree, contrary to the assumption that it has degree different from zero.
\end{proof}
In the sequel we will need also the following: 
\begin{lemma}\label{eq:degree}
Let $f: S^n \rightarrow S^n$ a continuous bijective map from the $n$-dimensional sphere to itself ($n\geq 1)$. Then $\mathrm{deg}(f)=\pm 1$. 
\end{lemma}
\begin{proof}
The map $f$ is a continuous bijective map from a compact space to a Hausdorff space, and therefore it is a homeomorphism. Now, a homeomorphism $f: S^n \rightarrow S^n$ has degree $\pm 1$ (see, for instance, \cite[Page 136]{Hatcher:01}).
\end{proof}

\subsection{Voronoi Diagrams and Power Diagrams}
\begin{figure}[thpb]
\centering  
    \mbox{
      \subfigure[A Voronoi Diagram.]
      {\label{fig:vorA}\scalebox{0.482}{\includegraphics{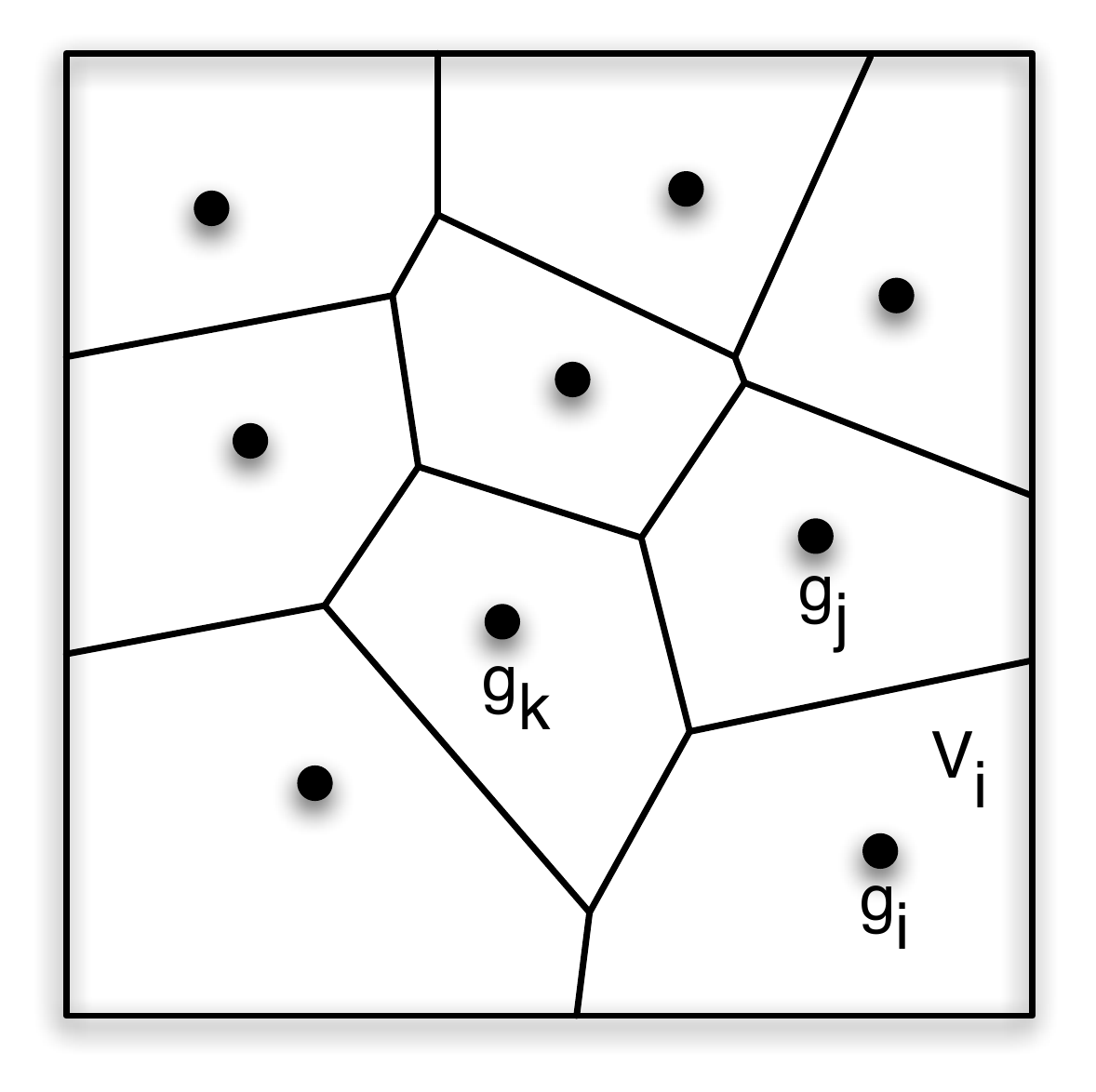}}}
\quad \quad \quad
      \subfigure[A power diagram. The weights $w_i$ are assumed positive. Circles represent the magnitudes of weights. Power generator $(g_2,w_2)$ has an empty cell. Power generator $(g_5, w_5)$ is outside its region of dominance.]
      {\label{fig:powB}\scalebox{0.25}{\includegraphics{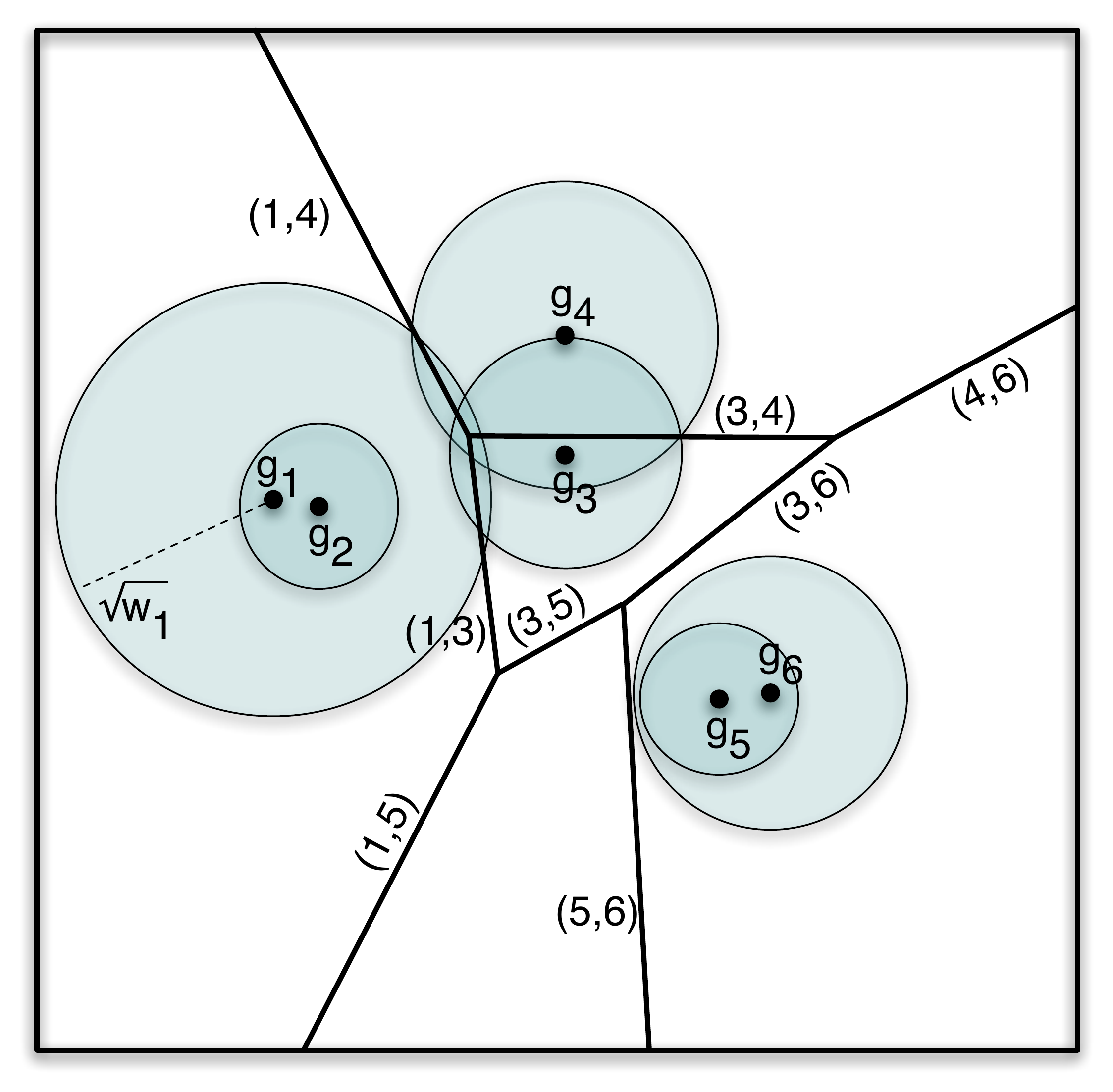}}}

} 
    \caption{Voronoi diagrams and power diagrams.}
    \label{fig:VorPow}
\end{figure}

We refer the reader to \cite{Okabe:00} and \cite{Imai.Iri.ea:85} for comprehensive
treatments, respectively, of Voronoi diagrams and power diagrams. Assume, first, that $G$ is an ordered set of \emph{distinct} points. The \emph{Voronoi diagram} $\mathcal V(G) = (V_1(G),\cdots,V_m(G))$ of $ A$ generated by points $(g_1,\cdots,g_m) $ is defined by
\begin{equation}
V_i(G) = \{x \in A |\, \, \|x  -g_i \| \leq \|x - g_j \|, \, \forall j \neq i, \, j \in I_m \}.
\end{equation}
We refer to $G$ as the set of \emph{generators} of $\mathcal V(G)$, and to $V_i(G)$  as the Voronoi cell or region of dominance of the $i$-th generator.
For $g_i,g_j\in G$, $i\neq j$, we define the \emph{bisector} between $g_i$ and $g_j$ as
$
b(g_i, g_j) = \{ x \in A |\, \,  \|x  -g_i \| = \|x - g_j \|\}.
$
The face $b(g_i, g_j)$ bisects the line segment joining $g_i$ and
$g_j$, and this line segment is orthogonal to the face (\emph{Perpendicular Bisector Property}). The bisector divides $A$ into two convex subsets, and leads to the definition of the set
$
D(g_i,g_j) = \{x \in A |\, \,  \|x  -g_i \| \leq \|x - g_j \| \}; 
$
we refer to $D(g_i,g_j)$ as the \emph{dominance region} of $g_i$ \emph{over} $g_j$. Then, the Voronoi partition $\mathcal V(G)$ can be equivalently defined as
$V_i(G) = \bigcap_{j\in I_m \setminus \{i\}} D(g_i,g_j).$
This second definition clearly shows that each Voronoi cell is a convex set. Indeed, a Voronoi diagram of $A$ is a convex partition of $A$ (see Fig. \ref{fig:vorA}). The Voronoi diagram of an ordered set of possibly \emph{coincident} points is not well-defined. We define
\begin{equation}\label{eq:scoinc}
\Gamma_{\text{coinc}} = \{(g_1, \cdots, g_m) \in A^m \,|\, g_i = g_j \text{ for some } i\neq j\in \{1,\cdots, m\}\}.  
\end{equation}

Assume, now, that each point $g_i \in G$ has assigned an individual
weight $w_i \in \reals$, $i\in I_m$; let $W=(w_1,\cdots,w_m)$. We define the power distance
\begin{equation}
d_P(x,g_i; w_i) \doteq \|x -g_i \|^2 - w_i.
\end{equation}

We refer to the pair $(g_i, w_i)$ as a \emph{power point}. We define $G_W = \Bigl((g_1,w_1),\cdots,(g_m,w_m) \Bigr)$. Two power points $(g_i, w_i)$ and $(g_j, w_j)$ are \emph{coincident} if  $g_i = g_j$ \emph{and} $w_i = w_j$. Assume, first, that $G_W$ is an ordered set of  \emph{distinct} power points. Similarly as before, the \emph{Power diagram} $\mathcal V(G_W) = (V_1(G_W),\cdots,V_m(G_W))$ of $ A$ generated by power points $\Bigl((g_1,w_1),\cdots,(g_m,w_m) \Bigr)$ is defined by
\begin{equation}
\begin{split}
V_i(G_W) = \{x \in A | \, \,  \|x -g_i \|^2 - w_i& \leq  \|x -g_j\|^2 - w_j, \, \forall j \neq i, \, j \in I_m \}.
\end{split}
\end{equation}
We refer to $G_W$ as the set of \emph{power generators} of $\mathcal
V(G_W)$, and to $V_i(G_W)$  as the power cell or region of dominance of the
$i$-th power generator; moreover we call $g_i$ and $w_i$, respectively, the
position and the weight of the power generator $(g_i, w_i)$. Notice that, when all weights
are the same, the power diagram of $A$ coincides with the Voronoi diagram of $A$. As before, power diagrams can be defined as intersection of convex sets; thus, a
power diagram is, as well, a convex partition of $A$. Indeed, power diagrams are the generalized Voronoi diagrams that have the strongest similarities to the original diagrams \cite{Aurenhammer:87}. There are some differences, though. First, a power cell might be empty. Second, $g_i$ might not be in its power cell (see Fig. \ref{fig:powB}). Finally, the bisector of $(g_i,w_i)$ and $(g_j,w_j)$,  $i\neq j$, is
\begin{equation}\label{eq:powerBisec}
b\Bigl((g_i,w_i), (g_j,w_j) \Bigr) = \{ x \in A | \, \, (g_j - g_i)^{\text{T}}x = \frac{1}{2}(\|g_j \|^2 - \| g_i\|^2 + w_i - w_j)\}.
\end{equation} 
Hence, $b\Bigl((g_i,w_i), (g_j,w_j) \Bigr) $ is a face  orthogonal to the line segment $\overline{g_i\,g_j}$ and passing through the point $g_{ij}^*$ given by
$$g_{ij}^* = \frac{\|g_j\|^2 - \| g_i \|^2
+w_i - w_j}{2\|g_j - g_i \|^2}(g_j - g_i); $$
this last property is crucial in the remaining of the paper: it means that, by changing weights, it is possible to arbitrarily move the bisector between the positions of two power generators, while still preserving the orthogonality constraint.

The power diagram of an ordered set of possibly \emph{coincident} power points is not well-defined. We define
\begin{equation}\label{eq:scoincPow}
\Gamma_{\text{coinc}} = \Bigl \{\Bigl ((g_1,w_1), \cdots, (g_m,w_m) \Bigr) \in (A \times \reals)^m \,|\, g_i = g_j \text{ and } w_i = w_j \text{ for some } i\neq j\in \{1,\ldots, m\} \Bigr\}.  
\end{equation}
Notice that we used the same symbol as in Eq. \eqref{eq:scoinc}: the meaning will be clear from the context. 


For simplicity, we will refer to $V_i(G)$ ($V_i(G_W)$) as $V_i$.  When the
two Voronoi (power) cells $V_i$ and $V_j$ are adjacent (i.e., they share a face), $g_i$ ($(g_i,w_i)$) is called a \emph{Voronoi} (\emph{power}) \emph{neighbor} of
$g_j$ ($(g_j,w_j)$), and vice-versa. The set of indices of the Voronoi (power) neighbors
of $g_i$ ($(g_i,w_i)$) is denoted by $N_i$. We also define the $(i,\,j)$-face as
$\Delta_{ij} \doteq V_i\cap V_j$. 


\subsection{Proximity Graphs and Spatially-Distributed Control Policies for Robotic Networks}

Next, we shall present some relevant concepts on proximity graph functions and spatially-distributed control policies; we refer the reader to \cite{JC-SM-FB:03p} for a more detailed discussion. A \emph{proximity graph function} $\mathcal G :\mathbb{F}(\reals^d) \to \mathbb{G}(\reals^d)$ associates to a point set $\mathcal P \in \mathbb{F}(\reals^d)$  an undirected graph with vertex set $\mathcal P$ and edge set $\mathcal E_{\mathcal  G}(\mathcal P)$, where $\mathcal E_{\mathcal  G}: \mathbb{F}(\reals^d) \mapsto \mathbb{F}(\reals^d \times \reals^d)$ has the property that $\mathcal E_{\mathcal  G}(\mathcal P) \subset \mathcal P\times \mathcal P \setminus \text{diag}(\mathcal P \times \mathcal P)$ for any $\mathcal P$ . Here, $\text{diag}(\mathcal P \times \mathcal P) = \{(p, p) \in \mathcal P \times \mathcal P | \, \, p\in \mathcal P\}$. In other words, the edge set of a proximity graph depends on the location of its vertices. To each proximity graph function, one can associate the \emph{set of neighbors map} $N_{\mathcal G}: \reals^d \times \mathbb{F}(\reals^d) \to \mathbb{F}(\reals^d)$, defined by
$$N_{\mathcal G}(p, \mathcal P) =  \{q\in \mathcal P | \,\, (p,q) \in \mathcal E_{\mathcal G}(\mathcal P \cup \{ p\})\}.$$
Two examples of proximity graph functions are:
\begin{enumerate}[(i)]
\item the \emph{Delaunay} graph $G \mapsto \mathcal G_{\text{V}}(G) = (G, \mathcal E_{\mathcal G_{\text{V}}}(G))$ has edge set 
$$\mathcal E_{\mathcal G_{\text{V}}}(G) = \{(g_i,g_j) \in G \times G \setminus \text{diag}(G \times G ) | \, \,V_i(G) \cap V_j(G) \neq \emptyset \},$$
where $V_i(G)$ is the $i$-th cell in the Voronoi diagram $\mathcal V(G)$; 
\item the \emph{power-Delaunay} graph $G_W \mapsto \mathcal G_{\text{P}}(G_W) = (G_W, \mathcal E_{\mathcal G_{\text{P}}}(G_W))$ has edge set 
$$\mathcal E_{\mathcal G_{\text{P}}}(G_W) = \Bigl \{\Bigl (g_i,w_i), (g_j,w_j)\Bigr) \in G_W \times G_W \setminus \text{diag}(G_W \times G_W ) | \, \, V_i(G_W) \cap V_j(G_W) \neq \emptyset \Bigr \},$$
where $V_i(G_W)$ is the $i$-th cell in the power diagram $\mathcal V(G_W)$.
\end{enumerate}

We are now in a position to discuss spatially-distributed algorithms for robotic networks in formal terms. Let $P(t) = (p_1(t), \ldots, p_m(t)) \in A^{m}$ be the ordered set of positions of $m$ agents in a robotic network. We denote the state of each agent $i \in I_m$ at time $t$ as $x_i(t) \in \reals^q$ ($x_i(t)$ can include the position of agent $i$ as well as other information). With a slight abuse of notation, let us denote by $I_i(t)$ the information available to agent $i$ at time $t$.
The information vector $I_i(t)$ is a subset of $x(t) \doteq (x_1(t),\ldots, x_m(t))$ of the form $I_i(t) = \{x_{i_1}(t),\ldots, x_{i_k}(t) \}$, $k\leq m$. We assume that $I_i(t)$ always includes $x_i(t)$. Let $\mathcal G$ be a proximity graph function defined over $P(t)$ (respectively over $P_W(t)$ if we also consider a weight $w_i(t)$ for each robot $i\in I_m$); we define $I_i^{N_{\mathcal G}}(t)$ as the information vector with the property $x_i(t) \in I_i^{N_{\mathcal G}}(t)$, and, for $j\neq i$,
$$x_j(t) \in I_i^{N_{\mathcal G}}(t) \Leftrightarrow p_j(t) \in N_{\mathcal G}(p_i(t), P(t)) \quad \biggl ( \, \Leftrightarrow (p_j(t), w_j(t)) \in N_{\mathcal G}\Bigl ((p_i(t),w_i(t)), P_W(t)\Bigr),\,\, \text{respectively} \biggr) .$$
In words, the information vector $I_i^{N_{\mathcal G}}(t)$ coincides with the states of the neighbors (as induced by $\mathcal G$) of agent $i$ together with the state of agent $i$ itself.

Let $\mu(t) = (\mu_1(I_1(t)), \ldots, \mu_m(I_m(t))$ be a feedback control policy for the robotic network. The policy $\mu$ is \emph{spatially distributed over} $\mathcal G$ if for each agent $i \in I_m$ and for all $t$
$$\mu_i(I_i(t)) =  \mu_i\Bigl (I_i^{N_{\mathcal G}}(t) \Bigr).$$
In other words, through information about its neighbors according to $\mathcal G$, each agent $i$ has sufficient information to compute the control $\mu_i$.


\section{Problem Formulation}
A total of $m$ identical mobile agents provide service in a compact, convex service
region $ A \subseteq \reals^d$.  Let $\lambda$ be a measure whose bounded support is $A$ (in other words, $\lambda$ is not zero only on $A$); for any set $ S $, we define
the \emph{workload} for region $S$ as $\lambda_{ S} \doteq \int_{ S}
\lambda (x) \, dx$. The measure $\lambda$ models service requests, and can represent, for example, the
density of customers over $A$, or, in a stochastic setting, their arrival
rate. Given the measure $\lambda$, a partition $\{A_i\}_i$ of the workspace $A$
is \emph{equitable} if $\lambda_{A_i}=\lambda_{A_j}$ for all $i,\,j \in
I_m$.

A \emph{partitioning policy} is an algorithm that, as a
function of the number $m$ of agents and, possibly, of their position and
other information, partitions a bounded workspace $A$ into $m$ openly disjoint subregions $A_i$, $i \in I_m$. Then, each agent $i$ is
assigned to subregion $A_i$, and each service request in $A_i$ receives service from the agent assigned to $A_i$.  We refer to subregion $A_i$ as the \emph{region of dominance} of agent $i$. Given a measure $\lambda$ and a partitioning policy, $m$ agents are in a \emph{convex equipartition configuration} with respect to $\lambda$ if the associated partition is equitable and convex.

In this paper we study the following problem: find a \emph{spatially-distributed} (in the sense discussed in Section \ref{sec:Background}) equitable partitioning policy that allows $m$ mobile agents to achieve a convex equipartition configuration (with respect to $\lambda$). Moreover, we consider the issue of convergence to equitable partitions with some special properties, e.g., where subregions have shapes similar to those of regular polygons.

\section{Leader-Election Policies}
\begin{figure}[thpb]
\centering  
    \mbox{
      \subfigure[Sweeping $A$]
      {\scalebox{0.4}{\includegraphics{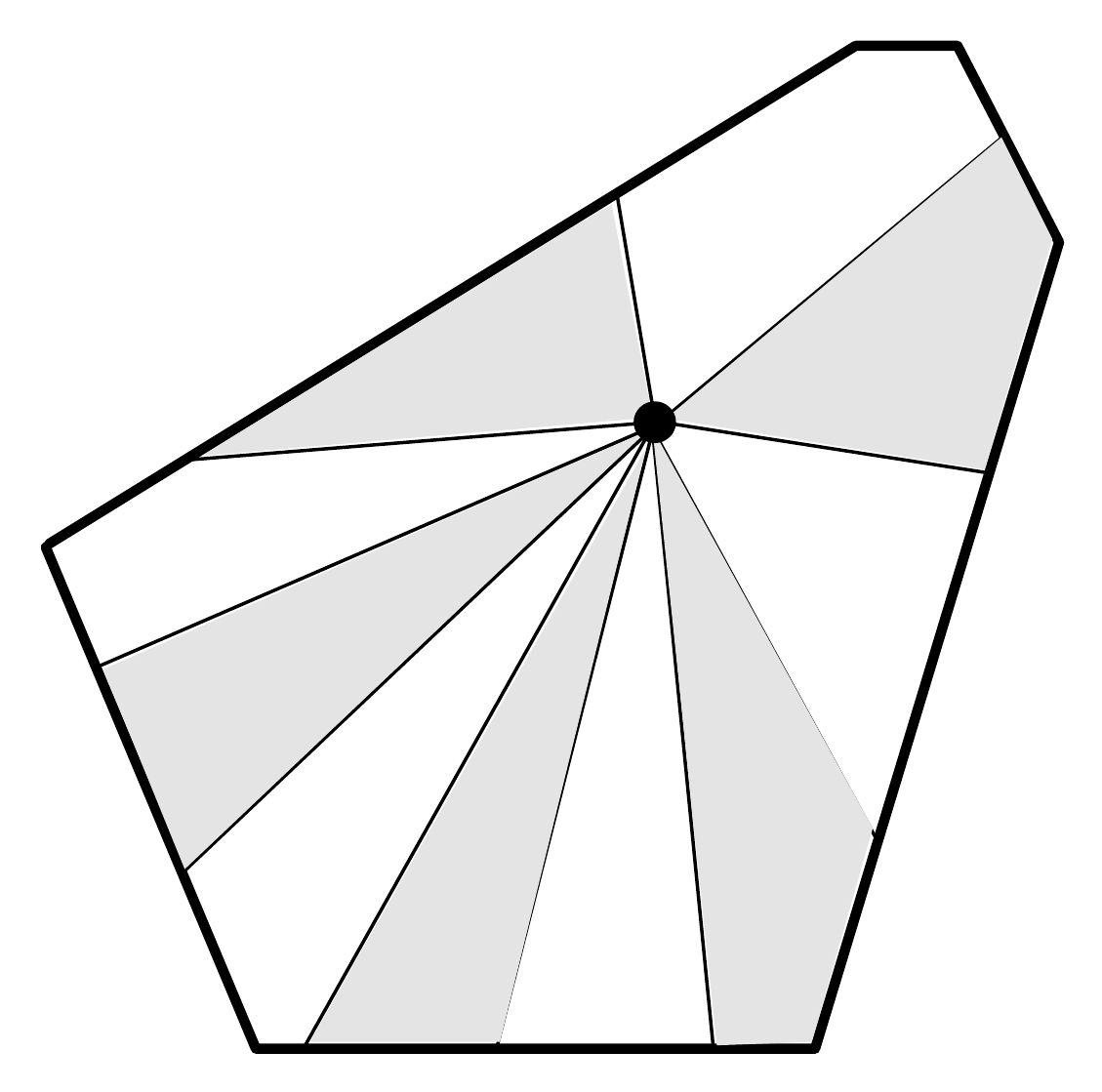}}}
\quad \quad \quad \quad
      \subfigure[Slicing $A$]
      {\scalebox{0.4}{\includegraphics{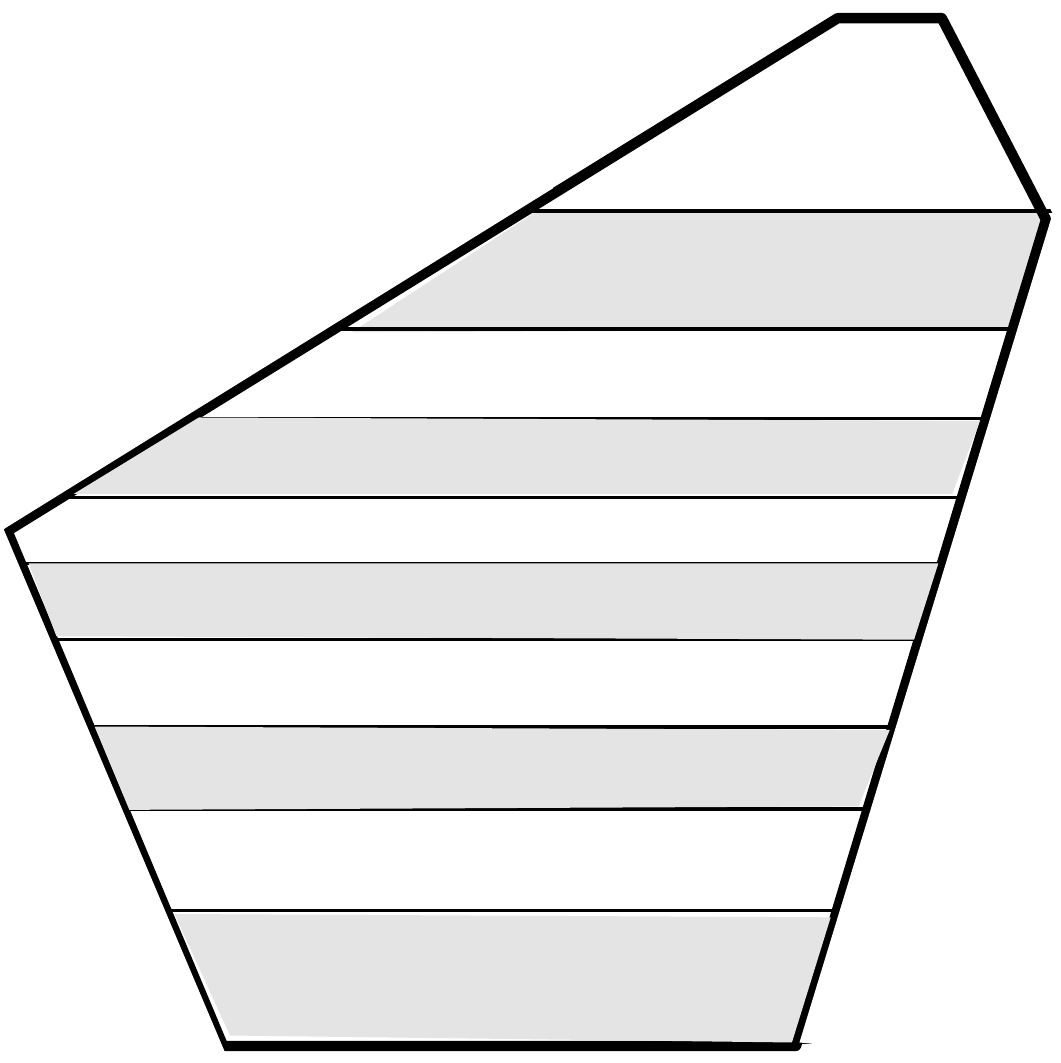}}}

} 
    \caption{Equitable partitions by sweeping and slicing (assuming a uniform measure $\lambda$).}
    \label{fig:sliceSweep}
\end{figure}

We first describe two simple algorithms that provide equitable partitions. A first possibility is to ``sweep" $A$ from a point in the interior of $A$ using an arbitrary starting ray until $\lambda_{A_1} = \lambda_A/m$, continuing the sweep until $\lambda_{A_2} = \lambda_A/m$, etc. A second possibility is to slice, in a similar fashion, $A$. The resulting equitable partitions are depicted in Fig. \ref{fig:sliceSweep} 

Then, a possible solution could be to (i) run a distributed leader election algorithm over the graph associated to some proximity graph function $\mathcal G$ (e.g., the Delaunay graph); (ii) let each agent send its state $x_i(t)$ to the leader; (iii) let the leader execute either the sweeping or the slicing algorithms described above; finally, (iv), let the leader broadcast subregion's assignments to all other agents. Such conceptually simple solution, however, can be impractical in scenarios where the density $\lambda$ changes over time, or agents can fail, since at every parameter's change a new time-consuming leader election is needed. Moreover, the sweeping and the slicing algorithms provide long and skinny subregions that are not suitable in most applications of interest (e.g., vehicle routing).

We now present spatially-distributed algorithms, based on the concept of power diagrams, that solve \emph{both issues} at once.

\section{Spatially-Distributed Gradient-Descent Law for Equitable Partitioning}

We start this section with an existence theorem for equitable power diagrams.

 \subsection{On the Existence of Equitable Power Diagrams}
As shown in the next theorem, an equitable power diagram (with respect to any $\lambda$) exists for \emph{any} vector of \emph{distinct} points $G= (g_1, \ldots, g_n)$ in $A$. 

\begin{theorem}\label{thm:homotopy}
Let $A$ be a bounded, connected domain in $\reals^d$, and $\lambda$ be a measure on $A$. Let $G = (g_1, \dots, g_n)$ be the positions of $n\geq 1$ distinct points in $A$. Then, there exist weights $w_i$, $i \in I_n$, such that the power points $\Bigl((g_1, w_1),\ldots,(g_n,w_n) \Bigr) $ generate a power diagram that is equitable with respect to $\lambda$. Moreover, given a vector of weights $w^*$ that yields an equitable power diagram, the set of all vectors of weights yielding an equitable power diagram is $\mathcal W^*_t \doteq \{ w^*+t[1,\ldots,1]\}$, with $t\in\reals$.
\end{theorem}
 
\begin{proof}
It is not restrictive to assume that $\lambda_A = 1$ (i.e., we normalize the measure of $A$), since $A$ is bounded. The strategy of the proof is to use a topological argument to force existence. 

First, we construct a weight space. Let $D=\mathrm{diameter}(A)$, and
consider the cube $\mathcal{C}:=[-D, D]^n$. This is the weight space and we
consider weight vectors $W$ taking value in $\mathcal{C}$. Second, consider
the standard $n$-\emph{simplex} of measures
$\lambda_{A_1},\ldots,\lambda_{A_n}$ (where $A_1,\dots,A_n$ are, as usual,
the regions of dominance). This can be realized in $\mathbb{R}^{n}$ as the
subset of defined by $\sum_{i=1}^n \lambda_{A_i}=1$ with the condition
$\lambda_{A_i} \geq 0$. Let us call this set ``the measure simplex $\mathcal{A}$" (notice that it is $(n-1)$-dimensional).
 
There is a map $f: \C \rightarrow \A$ associating, according to the power distance, a weight vector $W$ with the corresponding vector of measures $(\lambda_{A_1},\ldots,\lambda_{A_n})$. Since the points in $G$ are assumed to be distinct, this map is continuous.


We will now use induction on $n$, starting with the base case $n=3$ (the statement for $n=1$ and $n=2$ is trivially checked). We study in detail the case for $n=3$, where visualization is easier, in order to make the proof more transparent. When $n=3$, the weight space $\C$ is a three dimensional cube with vertices $v_0=[-D, -D, -D],$ $v_1=[D, -D, -D],$ $v_2=[-D, D, -D],$ 
$v_3=[-D, -D, D],$ $v_4=[D, -D, D],$ $v_5=[-D, D, D],$ $v_6=[D, D, -D]$ and $v_7 =[D, D, D]$. 
The measure simplex $\A$ is, instead, a triangle with vertices $u_1, u_2, u_3$ that correspond to the cases  $\lambda_{A_1}=1, \lambda_{A_2}=0, \lambda_{A_3}=0$,  $\lambda_{A_1}=0, \lambda_{A_2}=1, \lambda_{A_3}=0$, and $\lambda_{A_1}=0, \lambda_{A_2}=0, \lambda_{A_3}=1$, respectively. Moreover, call $e_1, e_2$ and $e_3$ the edges opposite the vertices $u_1, u_2, u_3$ respectively. The edges $e_i$ are, therefore, given by the condition $\lambda_{A_i} \in e_i \Leftrightarrow \lambda_{A_i}=0$. 

Let us return to the map $f: \C \rightarrow \A$ (now in the case of three generators). Observe that the map $f$ sends $v_0$ the unique point $p_0$ of $\A$ corresponding to the measures of usual Voronoi cells (since the weights are all equal). Call $l_1$ the edge $v_0v_1$; then, it is immediate to see that the image of $l_1$ through $f$ is a path $\gamma_1$ in $\A$ joining $p_0$ to $u_1$. Analogously, the image of $l_2=v_0v_2$ through $f$ is a path $\gamma_2$ in $\A$ joining $p_0 $ to $u_2$ and, finally,  the image of $l_3=v_0v_3$  through $f$ is a path $\gamma_3$ connecting $p_0$ to $u_3$ (see Fig. \ref{fig:proof-power}). 
Now, we observe that paths $\{\gamma_i\}_{\{i=1,2,3\}}$ do not intersect except in $p_0$. To prove this, start by observing that the image through $f$ of all the points on the main diagonal of the cube joining $v_0$ with $v_7$ is equal to a single point $p_0\in \A$. This is due to the fact that only the {\em differences} among weights change the vector $(\lambda_{A_1},\lambda_{A_2},\lambda_{A_3})$, i.e., if all weights are increased by the same quantity, the vector $(\lambda_{A_1},\lambda_{A_2},\lambda_{A_3})$ does not change. We will prove this in detail for the case of $n$-generators in the next few paragraphs. In particular, the image of the diagonal $v_0v_7$ is exactly the point for which the measures are those of a Voronoi partition. Now let us understand what are the ``fibers" of $f$, that is to say, the loci where $f$ is constant. Since the measures of the regions of dominance do not change if the differences among the weights are kept constant, then the fibers of $f$ in the weight space $\C$ are given by the equations $w_1-w_2=c_1$ \emph{and} $w_2-w_3=c_2$. Rearranging these equations, it is immediate to see that $w_1=w_3+c_1+c_2$, $w_2=w_3+c_2$ and $w_3=w_3$, therefore taking $w_3$ as parameter we see that the fibers of $f$ are  straight lines {\em parallel} to the main diagonal $v_0v_7$. On the weight space $\C$ let us define the following equivalence relation: $w \equiv w'$ if and only if they are on a line parallel to the main diagonal $v_0v_7$. Map $f: \C \rightarrow \A$ induces a continuous map (still called $f$ by abuse of notation) from $\C/\equiv $ to $\A$ having the same image. Let us identify $\C/\equiv $. It is easy to see that any line in the cube parallel to the main diagonal $v_0v_7$ is entirely determined by its intersection with the three faces $F_3=\{w_3=-D\}\cap\C$, $F_2=\{w_2=-D\}\cap\C$ and $F_1=\{w_1=-D\}\cap \C$. Call $\F$ the union of these faces. We therefore have a continuous map $f: \F \rightarrow \A$ that has the same image of original $f$; besides, in the few next paragraphs we will prove in general (i.e for the case with $n$ generators) that the induced map  $f: \F \rightarrow \A$ is {\em injective} by construction.
   
Observe that $\F $ is homeomorphic to $B^2$, the $2$-dimensional ball, like $\A$ itself. Up to homeomorphisms, therefore, the map $f : \F \rightarrow \A$ can be viewed as a map (again called $f$ by abuse of notation), $f: B^2 \rightarrow B^2$. Consider the closed loop $\alpha$ given by $v_2v_5$, $v_5v_3$, $v_3v_4$, $v_4v_1$, $v_1v_6$, $v_6v_2$ with this orientation. This loop is the boundary of $\F$ and we think of it also as the boundary of $B^2$. It is easy to see that $f$ maps $\alpha$ onto the boundary of $A$, and since $f$ is injective, the inverse image of any point in the boundary of $A$ is just one element of $\alpha$. Identifying the boundary of $\A$ with $S^1$ (up to homeomorphisms) and the loop $\alpha$ with $S^1$ (up to homeomorphisms) we have a bijective continuous map  $f_{S^1}: S^1 \rightarrow S^1$. By Lemma \eqref{eq:degree} $\mathrm{deg}(f)=\pm1$ and therefore $f$ is onto $\A$, using Theorem \eqref{eq:homotopy}. 

\begin{figure}[thpb]
\centering \scalebox{0.29} {\includegraphics{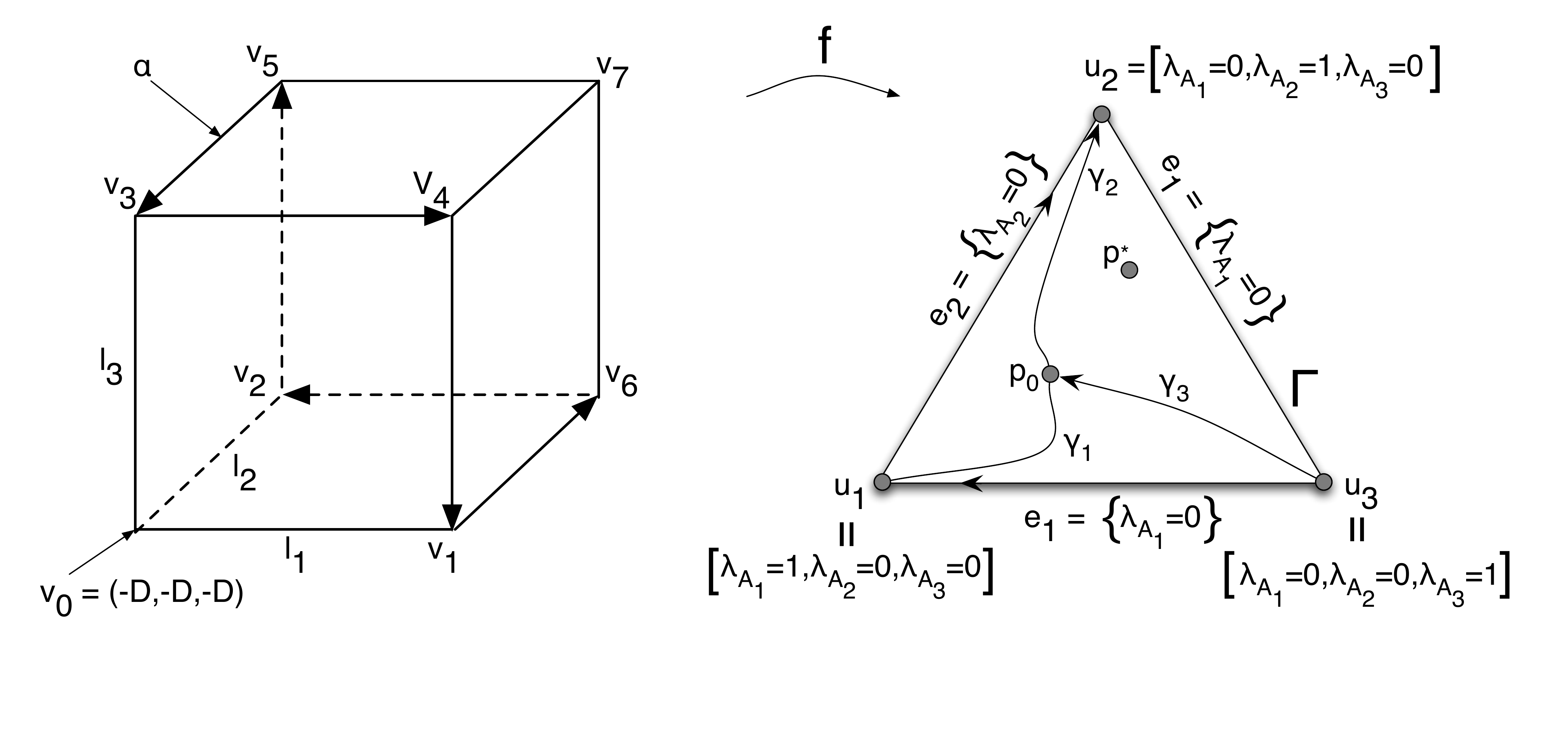}}
\caption{Construction used for the proof of existence of equitable power diagrams.} \label{fig:proof-power}
\end{figure}

Now we extend the same idea to the case of $n$ generators and we will use also induction on the number of agents. Therefore, we suppose that we have proved that the map $f$ is surjective for $n-1$ agents and we show how to use this to show that the map is surjective for $n$ agents.

If we have $n$ generators, the weight space is given by an $n$ dimensional cube $\mathcal{C}=[-D,D]^n$, in complete analogy with the case of $3$ generators. The $n$ simplex of the areas $\mathcal{A}$ is again defined as a $\{(\lambda_{A_1},\dots, \lambda_{A_n})\in\mathbb{R}^{n}\}$ such that $\lambda_{A_i}\geq 0$ for $i=1,\dots, n$ and $\sum_{i=1}^n \lambda_{A_i}=1$. Notice that $\mathcal{A}$ is homeomorphic to the $(n-1)$-dimensional ball $B^{n-1}$. As before we have a continuous map $f: \mathcal{C} \rightarrow \mathcal{A}$. It is easy to see that $f$ is constant on the sets of the form $W^*_t:=\{\{w^*+t(1,\dots, 1)\}\cap \mathcal{C}, \quad t\in \mathbb{R}\}$, that is whenever two sets of weights differ by a common quantity, they are mapped to the same point in $\mathcal{A}$. Moreover, fixing a point $Q\in \mathcal{A}$ we have that $f^{-1}(Q)$ is given just by a set of the form $W^*_t$ for a suitable $w^*$. Indeed, assume this is not the case, then the vector of measures $(\lambda_{A_1}, \dots, \lambda_{A_n})$ is obtained via $f$ using two sets of weights: $\mathcal{W}^1:=(w_1^1, \dots w_n^1)$ and $\mathcal{W}^2:=(w_1^2, \dots w_n^2)$, and $\mathcal{W}^1$ and $\mathcal{W}^2$ don't belong to the same $W^*_t$, namely it is is not possible to obtain $\mathcal{W}^2$ as $\mathcal{W}^1+t(1,\dots,1)$ for a suitable $t$. This means that the vector difference $\mathcal{W}^2-\mathcal{W}^1$ is not a multiple of $(1,\dots,1)$. Therefore, there exists a nonempty set of indexes $J$, such that $w^2_j-w^1_j\geq w^2_k-w^1_k$, whenever $j\in J$ and for all $k\in \{1,\dots n\}$ and such that the previous inequality is strict for at least one $k^*$.  Now among the indexes in $J$, there exists at least one of them, call it $j^*$ such that the agent $j^*$ is a neighbor of agent $k^*$, due to the fact that the domain $A$ is connected. But since $w^2_{j^*}-w^1_{j^*} >w^2_{k^*}-w^1_{k^*}$, and $w^2_{j^*}-w^1_{j^*}\geq w^2_k-w^1_k$ for all $k\in\{1,\dots, n\}$, this implies that  the measure $\lambda_{A_{j^*}}$ corresponding to the choice of weights $\mathcal{W}^2$ is strictly greater that $\lambda_{A_{j^*}}$ corresponding to the choice of weights  $\mathcal{W}^1$. This proves that $f^{-1}(Q)$ is given only by sets of the form $W^*_t$.

We introduce an equivalence relation on $\mathcal{C}$, declaring that two sets of weights $\mathcal{W}^1$ and $\mathcal{W}^2$ are equivalent if and only if they belong to the same $W^*_t$. Let us call $\equiv$ this equivalence relation. It is immediate to see that $f$ descends to a map, still called $f$ by abuse of notation, $f: \mathcal{C}/\equiv \rightarrow \mathcal{A}$ and that $f$ is now injective. It is easy also to identify $\mathcal{C}/\equiv$ with the union of the $(n-1)$-dimensional faces of $\mathcal{C}$ given by 
$\mathcal{F}=\cup_{i=1}^{n}(\mathcal{C}\cap\{w_i=-D\})$. In this way we get a continuous injective map $f:\mathcal{F} \rightarrow \mathcal{A}$ that has the same image as the original $f$. Notice also that $\mathcal{F}$ is homeomorphic to the closed $(n-1)$-dimensional ball, so up to homeomorphism $f$ can be viewed as a map $f: B^{n-1}\rightarrow B^{n-1}$.

We want to prove that the map $f_{\partial \mathcal{F}}$, given by the restriction of $f$ to $\partial \mathcal{F}$ is onto $\partial\mathcal{A}$. To see this, consider one of the $(n-2)$-dimensional faces $\partial \mathcal{A}_i$ of $\partial\mathcal{A}$, which is identified by the condition $\lambda_{A_i}=0$. Consider the face $F_i$ in $\mathcal{F}$, where $F_i$ is given by $F_i:=\mathcal{C}\cap\{w_i=-D\}$. We claim that the $S_i:=\partial F_i\cap \partial\mathcal{F}$ is mapped onto $\partial \mathcal{A}_i$ by $f$. Observe that the $S_i$ is described by the following equations $S_i=\cup_{j\neq i}(\{w_i=-D, w_j=D\}\cap \mathcal{F})$, so $S_i$ is exactly equivalent to a set of type $\mathcal{F}$ for {\em $n-1$ agents}. Moreover observe that $\partial \mathcal{A}_i$ can also be identified with the measure simplex for $n-1$ agents. By inductive hypothesis therefore, the map $f: S_i \rightarrow \partial \mathcal{A}_i$ is surjective, and therefore also the map $f_{\partial \mathcal{F}}$ is onto $\partial\mathcal{A}$. Since $f_{\partial \mathcal{F}}$ is a bijective continuos map among $(n-2)$-dimensional spheres, (up to homeomorphism), it has degree $\pm 1$ by Lemma \eqref{eq:degree}. Finally we conclude that $f$ is onto $\A$, using again Theorem \eqref{eq:homotopy}.

\end{proof}

Some remarks are in order.
\begin{remark}
The above theorem holds for any bounded, connected domain in $\reals^d$. Thus, the case of a compact, convex subset of $\reals^d$ is included as a special case.  Moreover, it holds for any measure $\lambda$ absolutely continuous with respect to the Lebesgue measure, and for \emph{any} vector of distinct points in $A$. 
\end{remark}
\begin{remark}\label{rem:generalExistence}
In the proof of the above theorem, we actually proved that for \emph{any} measure vector $\{\lambda_{A_i}\}_{i=1,\dots m}$ in $\A$, there exists a weight vector $w\in C$ realizing it through the map $f$. This could be useful in some applications.
\end{remark}
\begin{remark}
Since all vectors of weights in $\mathcal W$ yield exactly the \emph{same} power diagram, we conclude that the positions of the generators \emph{uniquely} induces an equitable power diagram.
\end{remark}

\subsection{State, Region of Dominance, and Locational Optimization}

The first step is to define the state for each agent $i$. We let $x_i(t)$ be the power generator $(g_i(t), w_i(t))\in A\times \reals$, where $g_i(t) = p_i(t)$ (i.e., the position of the power generator coincides with the position of the agent) \footnote{Henceforth, we assume that $A$ is a compact, convex subset of $\reals^2$.}. We, then, define the
region of dominance for agent $i$ as the power cell $ V_i = V_i(G_W)$,
where $G_W = \Bigl((g_1,w_1),\cdots,(g_m,w_m) \Bigr)$.
We refer
to the partition into regions of dominance induced by the set of
generators\footnote{For short, we will refer to a power generator simply as a generator.} $G_W$ as $\mathcal V(G_W)$. 

In light of Theorem~\ref{thm:homotopy}, the key idea is to enable the weights of the generators to follow a spatially-distributed
gradient descent law (while maintaining the positions of the generators \emph{fixed}) such that an equitable partition is reached. Assume, henceforth, that the positions of the generators are \emph{distinct}, i.e., $g_i \neq g_j$ for $i \neq j$. Define the set
\begin{equation}
S \doteq \Bigl \{( w_1,\ldots,w_m) \in \reals^m \, | \, \lambda_{V_i}>0 \, \, \, \forall i \in I_m \Bigr\}.
\end{equation}
Set $S$ contains all possible vectors of weights such that no region of dominance has measure equal to  \emph{zero}. 

We introduce the locational optimization function $H_V: S  \mapsto \reals_{>0}$:
\begin{equation}
 H_V( W)\doteq
 \sum_{i=1}^m  \Bigl(  \int_{ V_i(W)} \! \lambda(x) dx \Bigr)^{-1} 
 =  \sum_{i=1}^m  \lambda_{ V_i(W)} ^{-1}.
\end{equation}
where $W \doteq (w_1,\cdots,w_m)$.
\begin{lemma}
  A vector of weights that yields an equitable power diagram is a global minimum of $H_V$.
\end{lemma}
\begin{proof}
  Consider the relaxation of our minimization problem:
  \begin{equation*}
      \min_{x_1,\cdots,x_m} \sum_{i=1}^m x_i^{-1}\,; \quad
      \text{s.t. } \sum_{i=1}^m x_i = a>0, \quad x_i > 0, \; i \in
      I_m,
  \end{equation*}
  where the linear equality constraint follows from the fact that
  $\sum_{i=1}^m \int_{V_i(W)} \lambda(x) dx $ $ = \int_{ A} \lambda(x) \,dx
  \doteq a$ and where the constant $a$ is greater than zero since $\lambda$
  is a measure whose bounded support is $A$. By Lagrange multiplier
  arguments, it is immediate to show that the global minimum for the
  relaxation is $x_i = a/m$ for all $i$. Since Theorem~\ref{thm:homotopy}
  establishes that there exists a vector of weights in $S$ that yields an
  equitable power diagram, we conclude that this vector is a global minimum
  of $H_V$.
\end{proof}

\subsection{Smoothness and Gradient of $ H_{V}$}
We now analyze the smoothness properties of $H_V$. In the following, let $\gamma_{ij} \doteq \| g_j - g_i\|$.

\begin{theorem}\label{thrm:main} Assume that the positions of the generators are \emph{distinct}, i.e., $g_i \neq g_j$ for $i \neq j$. Given a measure $\lambda$, the function $H_V$ is continuously differentiable on $S$, where for each $i\in\{1,\dots,m\}$
\begin{equation}\label{eq:gradient}
\begin{split}
\frac{\partial  H_{ V}}{\partial \, w_i} &=   \sum _{j\in  N_i}  \frac{1}{2\gamma_{ij}}\Bigl (\frac{1}{ \lambda_{ V_j}^2} - \frac{1}{\lambda_{ V_i}^2}\Bigr ) \int_{\Delta_{ij}}\lambda(x)\, dx.
\end{split}
\end{equation}
Furthermore, the critical points of $H_V$ are vectors of weights that yield an equitable power diagram.
\end{theorem}\smallskip
\begin{proof}
By assumption, $g_i \neq g_j$ for $i \neq j$, thus the power diagram is well defined. Since the motion of a weight $w_i$ affects only power cell $V_i$ and its neighboring cells $V_j$ for $j \in   N_i$, we have
\begin{equation}
\label{eq:differentialH}
\frac{\partial  H_{V}}{\partial w_i} = -\frac{1}{\lambda_{ V_i}^2}\frac{\partial \lambda_{ V_i}}{\partial w_i} - \sum_{j\in  N_i}\frac{1}{ \lambda_{V_j}^2} \frac{\partial \lambda_{ V_j}}{\partial w_i} .
\end{equation}

Now, the result in Eq. (\ref{eq:divergence}) provides the means to analyze the variation of an integral function due to a domain change. Since the boundary of $ V_i$ satisfies $\partial  V_i = \cup_j \Delta_{ij} \cup  B_i$, where $\Delta_{ij} = \Delta_{ji}$ is the edge between $ V_i$ and $ V_j$, and $B_i$ is the boundary between $ V_i$ and $ A$ (if any, otherwise $B_i = \emptyset$), we have
\begin{equation}
\label{eq:powerDerivative1}
\begin{split}
\frac{\partial  \lambda_{ V_i}}{\partial w_i}   = \sum_{j \in  N_i} &\int_{\Delta_{ij}} \Bigl ( \frac{\partial x}{\partial w_i} \cdot n_{ij}(x) \Bigr ) \,  \lambda(x) \, dx + \underbrace{\int_{B_i} \Bigl ( \frac{\partial x}{\partial w_i} \cdot n_{ij}(x) \Bigr ) \,  \lambda(x) \, dx}_{=0},
\end{split}
\end{equation}
where we defined $n_{ij}$ as the unit normal to $\Delta_{ij}$ outward of $V_i$ (therefore we have $n_{ji} = -n_{ij}$).
The second term is trivially equal to zero if $B_i = \emptyset$; it is also equal to zero if $B_i \neq \emptyset$, since the integrand is zero almost everywhere. Similarly,
\begin{equation}
\label{eq:powerDerivative2}
\begin{split}
\frac{\partial  \lambda_{ V_j}}{\partial w_i}   =  &\int_{\Delta_{ij}} \Bigl ( \frac{\partial x}{\partial w_i} \cdot n_{ji}(x) \Bigr ) \,  \lambda(x) \, dx.
\end{split}
\end{equation}

To evaluate the scalar product between the boundary points and the unit outward normal to the border in Eq. (\ref{eq:powerDerivative1}) and in Eq. \eqref{eq:powerDerivative2}, we differentiate Eq. \eqref{eq:powerBisec} with respect to $w_i$ at every point $x \in \Delta_{ij}$; we get
\begin{equation}
\label{eq:bisectorDifferentiation}
 \frac{\partial x}{\partial w_i} \cdot (g_j - g_i) = \frac{1}{2}.
\end{equation}
From Eq. \eqref{eq:powerBisec} we have $n_{ij} = ~(g_j - ~g_i)~/\| g_j -
g_i\|$, and the desired explicit expressions for the scalar products in Eq.
\eqref{eq:powerDerivative1} and in Eq. \eqref{eq:powerDerivative2} follow
immediately (recalling that $n_{ji} = -n_{ij}$).

Collecting the above results, we get the partial derivative with respect to
$w_i$. The proof of the characterization of the critical points is an
immediate consequence of the expression for the gradient of $H_V$; we omit
it in the interest of brevity.
\end{proof}

\begin{remark}
The computation of the gradient in Theorem~\ref{thrm:main} is spatially-distributed over the power-Delaunay graph, since it
depends only on the location of the other agents with contiguous power
cells.
\end{remark}

\begin{example}[Gradient of $ H_{ V}$ for uniform measure] 
  The gradient of $ H_{ V}$ simplifies considerably when $\lambda$ is
  constant. In such case, it is straightforward to verify that (assuming
  that $\lambda$ is normalized)
\begin{equation}
\begin{split}
  \frac{\partial H_{ V}}{\partial \, w_i} &= \frac{1}{2 | A|}\sum _{j\in
    N_i}   \frac{\delta_{ij}}{\gamma_{ij}}\Bigl ( \frac{1}{| V_j|^2} - \frac{1}{| V_i|^2} \Bigr ),
\end{split}
\end{equation}
where $\delta_{ij}$ is the length of the border $\Delta_{ij}$.
\end{example}

\subsection{Spatially-Distributed Algorithm for Equitable Partitioning}
Consider the set $ U \doteq \Bigl \{( w_1,\ldots,w_m) \in \reals^m \, |
\, \sum_{i=1}^m w_i =0\Bigr\}$.  Since adding an identical value to every
weight leaves all power cells unchanged, there is \emph{no loss of
  generality} in restricting the weights to $U$; let $\Omega \doteq S \cap
U$. Assume the generators' weights obey a first order dynamical behavior
described by
$$ \dot{w}_i = u_i.$$
Consider $H_V$ an objective function to be minimized and impose that the weight $w_i$ follows the gradient descent given by \eqref{eq:gradient}. In more precise terms, we set up the following control law defined over the set $\Omega$
\begin{equation}\label{eq:vectorField}
u_i = -\frac{\partial H_V}{\partial w_i}(W),
\end{equation}
where we assume that the partition $\mathcal V(W) = \{V_1,\ldots,V_m \}$ is continuously updated. One can prove the following result.

\begin{theorem}\label{thrm:convergence} Assume that the positions of the generators are \emph{distinct}, i.e. $g_i \neq g_j$ for $i \neq j$. Consider the gradient vector field on $\Omega$ defined by equation \eqref{eq:vectorField}. Then generators' weights starting at $t=0$ at $W(0) \in \Omega$, and evolving under \eqref{eq:vectorField} remain in $\Omega$ and converge asymptotically to a critical point of $H_V$, i.e., to a vector of weights yielding an equitable power diagram.
\end{theorem}
\smallskip
\begin{proof}
We first prove that generators' weights evolving under \eqref{eq:vectorField} remain in $\Omega$ and converge asymptotically to the \emph{set} of critical points of $H_V$.
By assumption, $g_i \neq g_j$ for $i \neq j$, thus the power diagram is well defined. First, we prove that set $\Omega$ is positively invariant with respect to \eqref{eq:vectorField}. Recall that $\Omega = S \cap U$. Noticing that control law \eqref{eq:vectorField} is a gradient descent law, we have
$$\lambda_{V_{i}(W(t))}^{-1} \le H_{V}(W(t)) \le H_{V}(W(0)), \quad i \in I_m, \, \, t\geq 0.$$ 
Since the measures of the power cells depend continuously on the weights,
we conclude that the measures of all power cells will be bounded away from
zero; thus, the weights will belong to $S$ for all $t \geq 0$, that is, $W(t)\in S$ $\forall t\geq 0$. Moreover, the sum of the weights is invariant under control law \eqref{eq:vectorField}. Indeed, 
$$\frac{\partial \sum_{i=1}^m w_i}{\partial t} =  -\sum_{i=1}^m \frac{\partial H_V}{\partial w_i} =-\sum_{i=1}^m \sum _{j\in  N_i}  \frac{1}{2\gamma_{ij}}\Bigl (\frac{1}{ \lambda_{ V_j}^2} - \frac{1}{\lambda_{ V_i}^2}\Bigr ) \int_{\Delta_{ij}}\lambda(x)\, dx = 0,$$
since $\gamma_{ij} = \gamma_{ji}$, $\Delta_{ij} = \Delta_{ji}$, and $j \in
N_i \Leftrightarrow i \in N_j$. Thus, we have $W(t)\in U$ $\forall t\geq
0$. Since $W(t)\in S$ $\forall t\geq 0$ and $W(t)\in U$ $\forall t\geq 0$,
we conclude that $W(t)\in S\cap U = \Omega$, $\forall t\geq 0$, that is, set $\Omega$ is positively invariant.

Second, $H_V:\Omega \to \reals_{\geq 0}$ is clearly non-increasing along
system trajectories, that is, $\dot{H}_V \leq 0$ in $\Omega$. 

Third, all trajectories with initial conditions in $\Omega$ are bounded. Indeed, we have already shown that $\sum_{i=1}^m w_i = 0$ along system trajectories. This implies that weights remain within a bounded set: If, by contradiction, a weight could become arbitrarily positive large, another weight would become arbitrarily negative large (since the sum of weights is constant), and the measure of at least one power cell would vanish, which contradicts the fact that $S$ is positively invariant.

Finally, by Theorem~\ref{thrm:main}, $H_V$ is continuously differentiable in $\Omega$. Hence, by invoking the LaSalle invariance principle (see, for instance, \cite{FB-JC-SM:08}), under the descent flow \eqref{eq:vectorField}, weights will converge asymptotically to the \emph{set} of
critical points of $H_V$, that is not empty as confirmed by Theorem~\ref{thm:homotopy}. 

Indeed, by Theorem \ref{thm:homotopy}, we know that all vectors of weights yielding an equitable power diagram differ by a common translation. Thus, the largest invariant set of $H_V$ in $\Omega$ contains only one point. This implies that $\lim_{t \to \infty} W(t)$ exists and it is equal to a vector of weights that yields an equitable power diagram.

\end{proof}
Some remarks are in order.
\begin{remark}
By Theorem \ref{thrm:convergence}, for \emph{any} set of generators' distinct positions, convergence to an equitable power diagram is \emph{global} with respect to $\Omega$. Indeed, there is a very natural choice for the initial values of the weights. Assuming that at $t=0$ agents are in $A$ and in distinct positions, each agent initializes its weight to zero. Then, the initial partition is a Voronoi tessellation; since $\lambda$ is positive on $A$, each initial cell has nonzero measure, and therefore $W(0)\in \Omega$ (the sum of the initial weights is clearly zero).
\end{remark}
\begin{remark}
The partial derivative of $H_V$ with respect to the $i$-th weight only depends on the weights of the agents with neighboring power cells. Therefore, the gradient descent law \eqref{eq:vectorField} is indeed a \emph{spatially-distributed} control law over the power-Delaunay graph. We mention that, in a power diagram, each power generator has an average number of neighbors less than or equal to six \cite{Aurenhammer:87}; therefore, the computation of gradient \eqref{eq:vectorField} is scalable with the number of agents.
\end{remark}

\begin{remark}
The focus of this paper is on equitable partitions. Notice, however, that it is easy to extend the previous algorithm to obtain a spatially-distributed (again over the power-Delaunay graph) control law that provides \emph{any} desired measure vector $\{ \lambda_{A_i}\}$. In particular, assume that we desire a partition such that $$\lambda_{A_i} = \beta_i \lambda_A,$$
where $\beta_i\in(0,\, 1)$, $\sum_{j=1}^m \beta_j=1$.  If we redefine $H_V: S \mapsto \reals_{>0}$ as 
$$H_V(W) \doteq \sum_{i=1}^m\frac{\beta_i^2}{\lambda_{V_i(W)}},$$
then, following the same steps as before, it is possible to show  that, under control law
$$\dot w_i = -\frac{\partial H_V}{\partial w_i}(W)  =  \sum _{j\in  N_i}  \frac{1}{2\gamma_{ij}}\biggl (\frac{\beta^2_j}{ \lambda_{ V_j}^2} - \frac{\beta^2_i}{\lambda_{ V_i}^2}\biggr ) \int_{\Delta_{ij}}\lambda(x)\, dx,$$
the solution converges to a vector of weights that yields a power diagram with the property $\lambda_{A_i} = \beta_i\lambda_A$ (whose existence is guaranteed by Remark \ref{rem:generalExistence}).
\end{remark}

\begin{remark}
Define the set $\Gamma \doteq A^m\setminus \Gamma_{\text{coinc}}$ (where $\Gamma_{\text{coinc}}$ is defined in Eq. \eqref{eq:scoinc}). It is of interest to define and characterize the vector-valued function $W^* :\Gamma  \mapsto \Omega$ that associates to each non-degenerate vector of generators' positions a set of weights such that the corresponding power diagram is equitable. Precisely, we define $W^*(G)$ as $W^*(G) \doteq \lim_{t \to \infty} W(t),$ where $W(t) = (w_1(t),\ldots,w_m(t))$ is the solution of the differential equation, with \emph{fixed} vector of generators' positions $G$,
 $$\dot{w_i}(t) = -\frac{\partial H_V}{\partial w_i}(W(t)), \quad w_i(0) = 0, \quad i\in I_m.$$

By Theorem \ref{thrm:convergence}, $W^*(G)$ is a well-defined quantity
(since the limit exists) and corresponds  to an equitable power
diagram. The next lemma characterizes $W^*(G)$.

\begin{lemma} \label{lemma:Wcont}
The  function $W^*$ is continuous on $\Gamma$.
\end{lemma}
\proof See Appendix.
\endproof
\end{remark}

\subsection{On the Use of Power Diagrams}
A natural question that arises is whether a similar result can be obtained by using Voronoi diagrams (of which power diagrams are a generalization). The answer is positive if we constrain $\lambda$ to be constant over $A$, but it is negative for general measures $\lambda$, as we briefly discuss next.

Indeed, when $\lambda$ is \emph{constant} over $A$, an equitable Voronoi diagram always exists. We prove this result in a slightly more general set-up. 
\begin{definition}[Unimodal Property]
Let $ A \subset \reals^d$ be a bounded, measurable set (not necessarily convex). We say that $A$ enjoys the Unimodal Property if there exists a unit vector $v \in \reals^d$ such that the following holds. For each $s \in \reals$, define the slice
$ A^s \doteq \{x \in  A, \quad  v \cdot x = s \}, $
and call 
$\psi (s) \doteq m_{d-1}( A^s) $
the $(d-1)$-dimensional Lebesgue measure of the slice. Then, the function $\psi$ is unimodal. In other words, $\psi$ attains its global maximum at a point $\bar{s}$, is increasing on $(-\infty, \bar{s}]$, and decreasing on $[\bar{s},\infty)$.
\end{definition} 

The Unimodal Property (notice that every compact, convex set enjoys such
property) turns out to be a sufficient condition for the existence of
equitable Voronoi diagrams for bounded measurable sets (with respect to
constant $\lambda$).

\begin{theorem}\label{th:Voronoi} If $ A \subset \reals^d$ is any bounded measurable set satisfying the Unimodal Property and $\lambda$ is constant on $A$, then for every $m \geq 1$ there exists an equitable Voronoi diagram with $m$ (Voronoi) generators.
\end{theorem}
\proof
See Appendix.
\endproof

Then, an equitable Voronoi diagram can be achieved by using a gradient descent law conceptually similar to the one discussed previously (the details are presented in \cite{Pavone.Frazzoli.ea:07}). We emphasize that the existence result on equitable Voronoi diagrams seems to be new in the rich literature on Voronoi tessellations.

While an equitable Voronoi diagram always exists when $\lambda$ is \emph{constant} over $A$, in general, for non-constant $\lambda$, an equitable Voronoi diagram fails to exist, as the following counterexample shows.
\begin{figure}[thpb]
\centering \scalebox{0.48} {\includegraphics{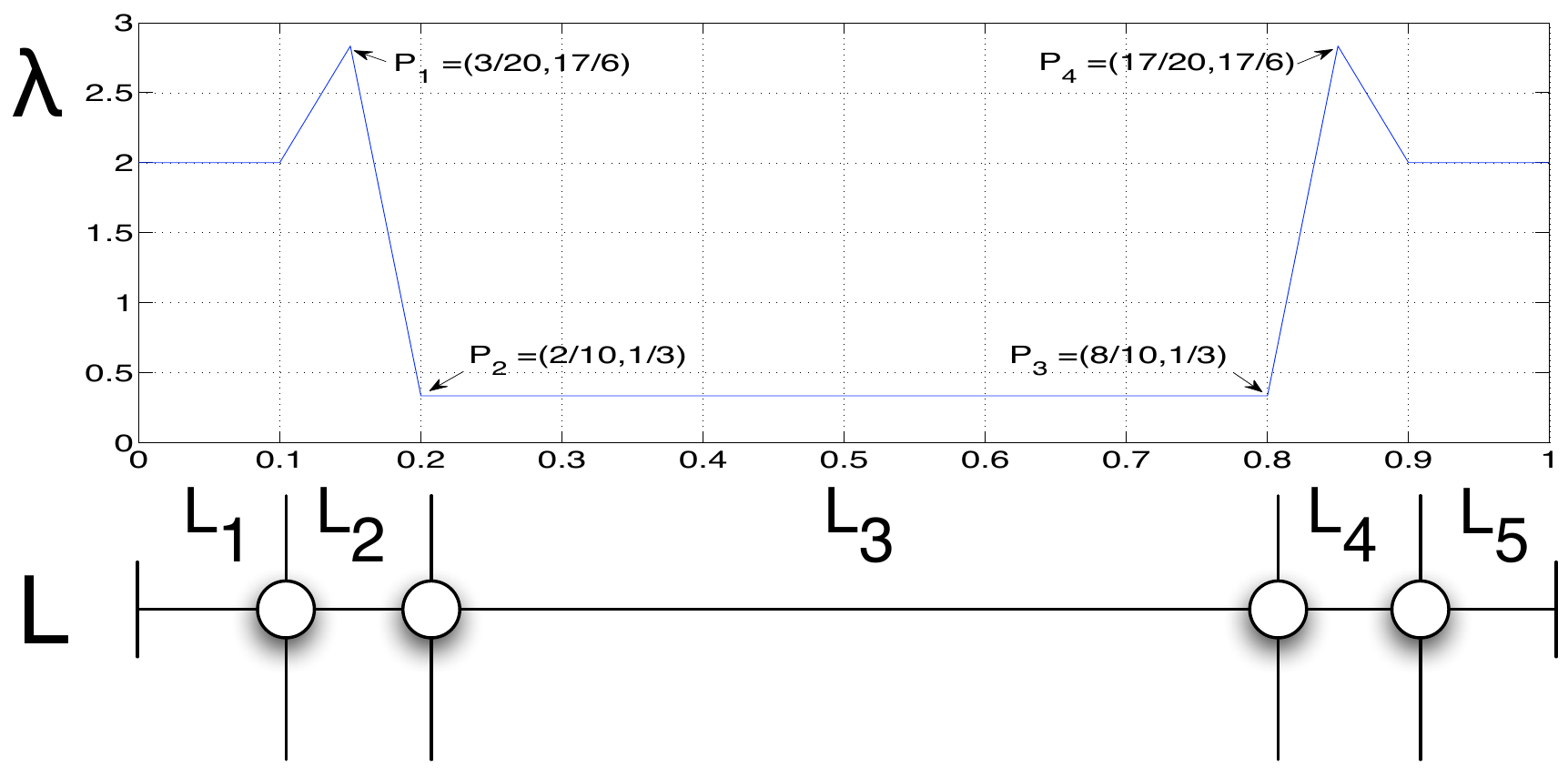}}
\caption{Example of non-existence of an equitable Voronoi diagram on a line. The above tessellation is an equitable partition, but not a Voronoi diagram.} \label{fig:counter}
\end{figure}
\begin{example}[Existence problem on a line]
Consider a one-dimensional Voronoi diagram. In this case a Voronoi cell is a half line or a line segment (called a Voronoi line), and Voronoi vertices are end points of Voronoi lines. It is easy to notice that the boundary point between two adjacent Voronoi lines is the mid-point of the generators of those Voronoi lines. Consider the measure $\lambda$ in Fig. \ref{fig:counter}, whose support is the interval $[0,\,1]$. Assume $m=5$. Let $b_i$ ($i=1,\ldots,4$) be the position of the $i$-th rightmost boundary point and $g_i$ be the position of the $i$-th rightmost generator ($i=1,\ldots,5$). It is easy to verify that the only admissible configuration for the boundary points in order to obtain an equitable Voronoi diagram is the one depicted in Fig. \ref{fig:counter}. Now, by the Perpendicular Bisector Property, we require:
\begin{displaymath}
\left\{ \begin{array}{l}
g_3 - b_2 = b_2 - g_2 \\
g_4 - b_3 = b_3  - g_3
\end{array} \right.
\end{displaymath}
Therefore, we would require $g_4  -g_2 = 2(b_3 - b_2) = 1.2$; this is impossible, since $g_2 \in [0.1,\,0.2]$ and $g_4 \in [0.8,\,0.9]$.
\end{example}

\section{Distributed Algorithms for Equitable Partitions with Special Properties}
The gradient descent law \eqref{eq:vectorField}, although effective in providing convex equitable partitions, can yield long and ``skinny" subregions. In this section, we provide spatially-distributed algorithms to obtain convex equitable partitions with special properties. The key idea is that, to obtain an equitable power diagram, changing the weights, while maintaining the generators \emph{fixed}, is sufficient. Thus, we can use the degrees of freedom given by the positions of the generators to optimize \emph{secondary} cost functionals. Specifically, we now assume that both generators' weights and their positions obey a first order dynamical behavior 
\begin{equation*}
\left \{ \begin{array}{l}
\dot w_i = u_i^w,\\
\dot{g}_i = u_i^g.
\end{array} \right.
\end{equation*}

Define the set
\begin{equation}
\tilde S \doteq \Bigl \{\Bigl( (g_1, w_1),\ldots,(g_m, w_m) \Bigr) \in (A \times \reals)^m \, | \, g_i \neq g_j \, \text{for all} \, i \neq j, \text{and} \, \lambda_{V_i}>0 \, \, \, \forall i \in I_m \Bigr\}.
\end{equation} 
The \emph{primary} objective is to achieve a convex equitable partition and is captured, similarly as before, by the cost function $\tilde H_V : \tilde S \mapsto \reals_{>0}$
$$\tilde H_V(G_W) \doteq \sum_{i=1}^m \lambda_{V_i(G_W)}^{-1}.$$

We have the following
\begin{theorem}\label{thrm:main2}Given a measure $\lambda$, the function $\tilde H_V$ is continuously differentiable on $\tilde S$, where for each $i\in\{1,\dots,m\}$
\begin{equation}\label{eq:gradientHtilde}
\begin{split}
\frac{\partial  \tilde H_{ V}}{\partial \, g_i} &= \sum _{j\in  N_i} \Bigl (\frac{1}{ \lambda_{ V_j}^2} - \frac{1}{\lambda_{ V_i}^2}\Bigr ) \int_{\Delta_{ij}} \frac{(x - g_i)}{\gamma_{ij}}\lambda(x)\, dx, \\ 
\frac{\partial  \tilde H_{ V}}{\partial \, w_i} &=   \sum _{j\in  N_i} \Bigl (\frac{1}{ \lambda_{ V_j}^2} - \frac{1}{\lambda_{ V_i}^2}\Bigr ) \int_{\Delta_{ij}} \frac{1}{2\gamma_{ij}}\lambda(x)\, dx.
\end{split}
\end{equation}
Furthermore, the critical points of $\tilde H_V$ are generators' positions and
weights with the property that all power cells have measure
equal to $\lambda_A/m$.
\end{theorem}\smallskip
\begin{proof} The proof of this theorem is very similar to the proof of Theorem \ref{thrm:main}; we omit it in the interest of brevity (the derivation of the partial derivative $\frac{\partial  \tilde H_{ V}}{\partial \, g_i}$ can be found in \cite{Pavone.Frazzoli.ea:CDC08}). 
\end{proof}
Notice that the computation of the gradient in Theorem~\ref{thrm:main2} is spatially distributed over the power-Delaunay graph. For short, we define the vectors $v_{\pm \partial \tilde H_i} \doteq \pm \frac{\partial \tilde H_V}{\partial g_i}$.

Three possible \emph{secondary} objectives are discussed in the remainder of this section.

\subsection{Obtaining Power Diagrams Similar to Centroidal Power Diagrams}\label{sec:powcent}
Define the \emph{mass} and the \emph{centroid} of the power cell $V_i$, $i \in I_m$, as
$$M_{V_i} = \int_{V_i} \lambda(x)\, dx , \quad C_{V_i} = \frac{1}{M_{V_i}}\int_{V_i} x\lambda(x)\, dx.$$
In this section we are interested in the situation where $g_i = C_{V_i}$, for all $i \in I_m$. We call such a power diagram a \emph{centroidal power diagram}. The main motivation to study centroidal power diagrams is that, as it will be discussed in Section \ref{sec:isoDefect}, the corresponding cells, under certain conditions, are \emph{similar} in shape to regular polygons. 

 
A natural candidate control law to try to obtain a centroidal and equitable power diagram (or at least a \emph{good} approximation of it) is to let the positions of the generators move toward the centroids of the corresponding regions of dominance, when this motion does not increase the disagreement between the measures of the cells (i.e., it does not make the time derivative of $\tilde H_V$ positive). 

First we introduce a $C^{\infty}$ saturation function as follows:
\begin{equation*}\label{eqcutoff}
\Theta(x)\doteq
\Biggl \{
\begin{array}{ll}
0  &\text{for $x\leq0$ } , \\
 \exp\Bigl (-\frac{1}{(\beta x)^2}\Bigr) &\text{for $x>0 , \quad \beta \in \reals_{>0}$}.
\end{array}
\end{equation*}
Moreover, we define the vector
$v_{C,g_i} \doteq  C_{V_i} - g_i$. Then, we set up the following control law defined over the set $\tilde S$, where we assume that the partition $\mathcal V(G_W) = \{V_1,\ldots,V_m \}$ is continuously updated,
\begin{equation}\label{eq:vectorField2}
\begin{split}
\dot{w}_i &= -\frac{\partial \tilde H_V}{\partial w_i}, \\ 
\dot{g}_i&= 
\frac{2}{\pi} \arctan \left[ \frac{\| v_{-\partial \tilde H_i} \|^2}{\alpha}\right] \, \, \Theta(v_{C,g_i} \cdot v_{-\partial \tilde H_i}) \, \,v_{C,g_i}, \quad \quad \alpha \in \reals_{>0}.
\end{split}
\end{equation}

In other words, $g_i$ moves toward the centroid of its cell \emph{if and only if} this motion is compatible with the minimization of $\tilde H_V$. In particular, the term $\arctan\Bigl (\| v_{-\partial \tilde H_i} \|^2/\alpha \Bigr)$ is needed to make the right hand side of \eqref{eq:vectorField2} $C^1$, while the term $\Theta(v_{C,g_i} \cdot v_{-\partial \tilde H_i})$ is needed to make the right hand side of \eqref{eq:vectorField2}  compatible with the minimization of $\tilde H_V$. To prove that the vector field is $C^1$ it is simply sufficient to observe that it is the composition and product of $C^1$ functions. Furthermore, the compatibility condition of the flow \eqref{eq:vectorField2} with the minimization of $\tilde H_V$ stems from the fact that $\dot{g}_i=0$ as long as $v_{C,g_i} \cdot v_{-\partial \tilde H_i}\leq 0$, due to the presence of $\Theta(\cdot)$. Notice that the computation of the right hand side of \eqref{eq:vectorField2} is spatially distributed over the power-Delaunay graph.

As in many algorithms that involve the update of generators of Voronoi diagrams, it is possible that under control law \eqref{eq:vectorField2} there exists a time $t
^*$ and $i, j \in I_m$ such that $g_i(t^*) = g_j(t^*)$. In such a case, either the power diagram is not defined (when $w_i(t^*) = w_j(t^*)$), or there is an empty cell ($w_i(t^*) \neq w_j(t^*)$), and there is no obvious way to specify the behavior of control law  \eqref{eq:vectorField2} for these singularity points. Then, to make the set $\tilde S$ positively invariant, we have to slightly modify the update equation for the positions of the generators. The idea is to stop the positions of two generators when they are \emph{close} and on a \emph{collision course}.

\begin{figure}[thpb]
\centering \scalebox{0.45} {\includegraphics{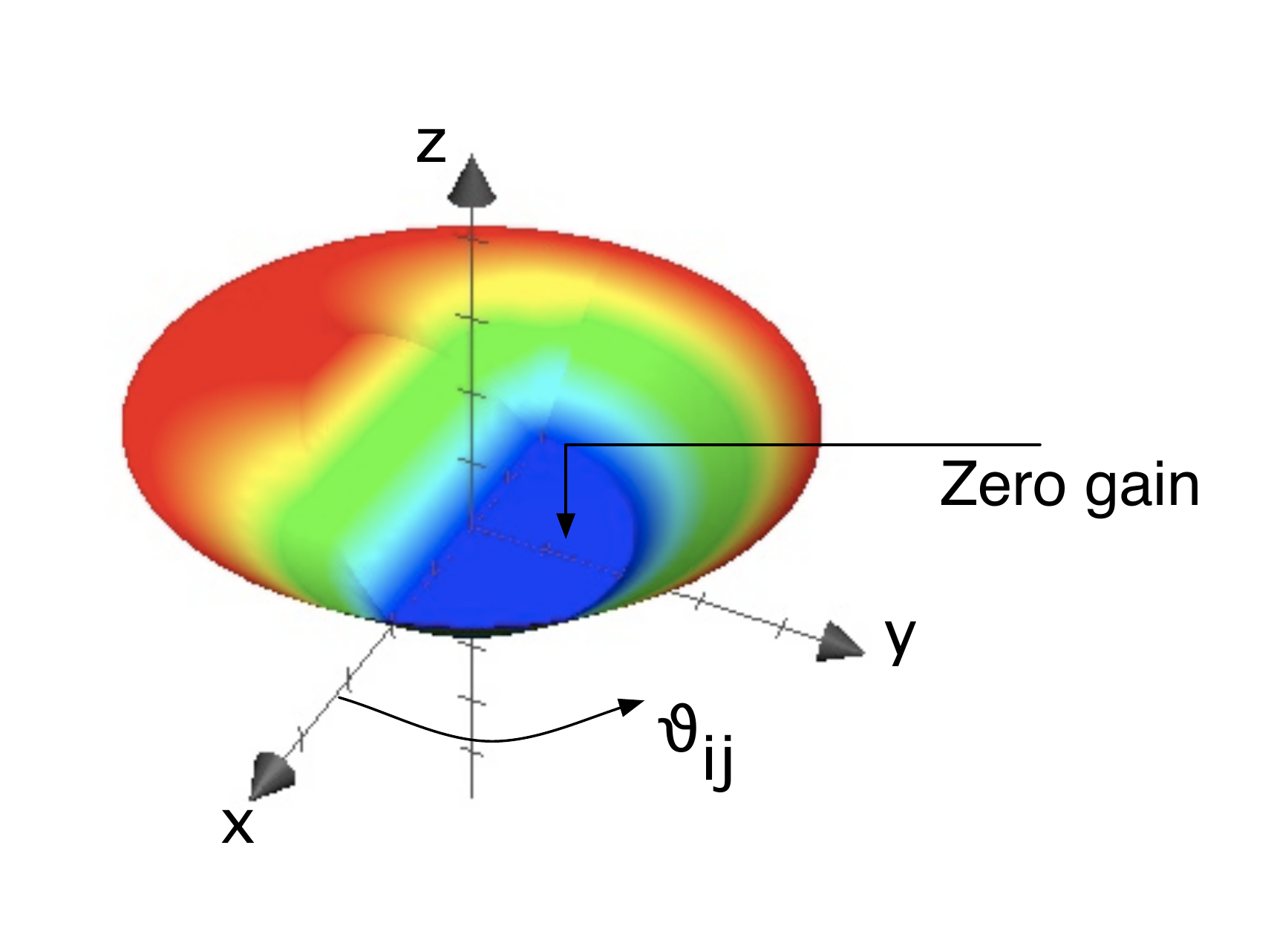}}
\caption{Gain function used to avoid that the positions of two power generators can coincide. }\label{fig:gain}
\end{figure}

Define, for $\Delta \in \reals_{>0}$, the set
$$M_i(G,\Delta) \doteq \{g_j \in G \, | \, \| g_i - g_j\|\leq \Delta,\, g_j\neq g_i  \} .$$
In other words, $M_i$ is the set of generators' positions within an (Euclidean) distance $\Delta$ from $g_i$. For $\delta \in \reals_{>0}$, $\delta<\Delta$, define the gain function $\psi(\rho,\vartheta):[0,\Delta] \times [0,2\pi] \mapsto \reals_{\geq 0}$ (see Fig. \ref{fig:gain}):
\begin{equation}\label{eq:psi}
\psi(\rho,\vartheta) = \left\{ \begin{array}{lll}
\frac{\rho - \delta}{\Delta -\delta} \quad & \textrm{if } \delta<\rho \leq \Delta &\text{ and } 0\leq \vartheta <\pi,\\
\frac{\rho - \delta}{\Delta -\delta}(1+\sin\vartheta) - \sin\vartheta\quad & \textrm{if } \delta<\rho \leq \Delta &\text{ and } \pi \leq  \vartheta \leq 2\pi,\\
0 & \textrm{if } \rho \leq \delta &\text{ and } 0 \leq  \vartheta <\pi,\\
-\frac{\rho}{\delta}\sin\vartheta & \textrm{if } \rho \leq \delta &\text{ and } \pi \leq  \vartheta \leq 2\pi,\\
\end{array} \right.
\end{equation}
It is easy to see that $\psi(\cdot)$ is a continuous function on $[0,\Delta] \times [0,2\pi]$  and it is globally Lipschitz there.
Function $\psi(\cdot)$ has the following motivation. Let $\rho$ be equal to $\|g_j-g_i\|$, 
and let $v_x$ be a vector such that the tern $\{v_x, (g_j-g_i), v_x \times (g_j-g_i)\}$ is an orthogonal basis of $\mathbb{R}^3$, co-orientied with the standard basis. In the Figure \ref{fig:gain}, $v_x$ corresponds to the $x$ axis and $g_j-g_i$ corresponds to the $y$ axis.
Then the angle $\vartheta_{ij}$ is the angle between $v_x$ and $v_{C,g_i}$, where $0\leq \vartheta_{ij}\leq 2\pi$. 
If  $\rho \leq \delta\text{ and } 0\leq \vartheta_{ij} <\pi$, then $g_i$ is \emph{close} to $g_j$ and it is on a \emph{collision course}, thus we set the gain to zero. Similar considerations hold for the other three cases; for example, if $\rho \leq \delta\text{ and } \pi < \vartheta_{ij} < 2\pi$, the generators are \emph{close}, but they are not on a collision course, thus the gain is positive.

Thus, we modify control law \eqref{eq:vectorField2} as follows:
\begin{equation}\label{eq:vectorField3}
\begin{split}
\dot{w}_i &= -\frac{\partial \tilde H_V}{\partial w_i}  \doteq u_i^{\text{cent,w}},\\
\dot{g}_i&=\frac{2}{\pi} \arctan \left[ \frac{\| v_{-\partial \tilde H_i} \|^2}{\alpha}\right]  \, \,  \Theta(v_{C,g_i} \cdot v_{-\partial \tilde H_i}) \, \,v_{C,g_i} \,\,  \prod_{g_j \in M_i(G,\Delta)}\psi\Bigl (\|g_j-g_i\|,\, \,\vartheta_{ij} \Bigr)  \doteq u_i^{\text{cent,g}}.
 \end{split}
\end{equation}
If $M_i(G,\Delta)$ is the empty set, then we have an empty product, whose numerical value is $1$. Notice that the right hand side of \eqref{eq:vectorField3} is Lipschitz continuous, since it is a product of $C^1$ functions and Lipschitz continuous functions, and it  can be still computed in a spatially-distributed way (in fact, it only depends on generators that are neighbors in the power diagram, and whose positions are within a distance $\Delta$). One can prove the following result.

\begin{theorem}\label{thrm:convergenceCentroidal} Consider the vector field on $\tilde S$ defined by equation \eqref{eq:vectorField3}. Then generators' positions and weights starting at $t=0$ at $G_W(0) \in \tilde S$, and evolving under \eqref{eq:vectorField3} remain in $\tilde S$ and converge asymptotically to the set of critical points of the primary objective function $\tilde H_V$ (i.e., to the set of vectors of generators' positions and weights that yield an equitable power diagram).
\end{theorem}
\smallskip
\begin{proof}
The proof is virtually identical to the one of Theorem \ref{thrm:convergence}, and we omit it in the interest of brevity. We only notice that $\tilde H_V$ is non-increasing along system trajectories  
$$\dot{\tilde H}_V = \sum_{i=1}^m \frac{\partial \tilde H_V}{\partial g_i}\cdot \dot g_i + \frac{\partial \tilde H_V}{\partial w_i}\dot w_i  = \sum_{i=1}^m \underbrace{\frac{\partial \tilde H_V}{\partial g_i}\cdot \dot g_i}_{\leq 0} - \Bigl (\frac{\partial \tilde H_V}{\partial w_i}\Bigr)^2 \leq 0.$$
Moreover, the components of the vector field \eqref{eq:vectorField3} for the position of each generator are either zero or point toward $A$ (since the centroid of a cell must be within $A$); therefore, each generator will remain within the compact set $A$.
\end{proof}

\subsection{Obtaining Power Diagrams ``Close" to Voronoi Diagrams}
In some applications it could be preferable to have power diagrams as \emph{close} as possible to Voronoi diagrams. This issue is of particular interest for the setting with non-uniform density, when an equitable Voronoi diagram could fail to exist. The objective of obtaining a power diagram \emph{close} to a Voronoi diagram can be translated in the minimization of the function $K : \reals^m \to \reals_{\geq 0}$:
$$K(W) \doteq \frac{1}{2} \sum_{i=1}^m w_i^2;$$
when $w_i = 0$ for all $i \in I_m$, we have $K=0$ and the corresponding power diagram coincides with a Voronoi diagram. To include the minimization of the secondary objective $K$, it is natural to consider, instead of \eqref{eq:vectorField}, the following update law for the weights:
\begin{equation}\label{eq:voronoiDefectFieldTemp}
\dot w_i = -\frac{\partial {H}_V}{\partial w_i} - \frac{\partial K}{\partial w_i} = -  \frac{\partial {H}_V}{\partial w_i} - w_i.
\end{equation}
However, $H_V$ is no longer a valid Lyapunov function for system \eqref{eq:voronoiDefectFieldTemp}. The idea, then, is to let the positions of the generators move so that
$\frac{\partial \tilde{H}_V}{\partial g_i}\cdot \dot g_i -  \frac{\partial \tilde{H}_V}{\partial w_i}\frac{\partial K}{\partial w_i} = 0$.
In other words, the dynamics of generators' positions is used to compensate the effect of the term $-w_i$ (present in the weights' dynamics) on the time derivative of $\tilde{H}_V$.

Thus, we set up the following control law, with $\varepsilon_1, \, \varepsilon_2$ and $\varepsilon_3$ positive \emph{small} constants, $\varepsilon_2>\varepsilon_1$, 
\begin{equation}\label{eq:vectorFieldVoronoi}
\begin{split}
\dot{w}_i &= -\frac{\partial \tilde{H}_V}{\partial w_i} - w_i \text{sat}_{\varepsilon_1,\varepsilon_2} \Bigl ( \|v_{\partial \tilde{H}_i}\| \Bigr) \,\text{sat}_{0,\varepsilon_3} \Bigl (\text{dist}(g_i,\partial V_i)\Bigr),\\
\dot{g}_i &=w_i\frac{\partial \tilde{H}_V}{\partial w_i} \frac{v_{\partial \tilde{H}_i}}{\|\ v_{\partial \tilde{H}_i} \|^2}\,\text{sat}_{\varepsilon_1,\varepsilon_2} \Bigl ( \|v_{\partial \tilde{H}_i}\| \Bigr) \,\text{sat}_{0,\varepsilon_3} \Bigl (\text{dist}(g_i,\partial V_i)\Bigr).
\end{split}
\end{equation}
The gain $\text{sat}_{\varepsilon_1,\varepsilon_2}\Bigl ( \|v_{\partial \tilde H_i}\| \Bigr)$ is needed to make the right hand side of \eqref{eq:vectorFieldVoronoi} Lipschitz continuous, while the gain  $\text{sat}_{0,\varepsilon_3} \Bigl (\text{dist}(g_i,\partial V_i)\Bigr)$ avoids that generators' positions can leave the workspace. Notice that  the computation of the right hand side of \eqref{eq:vectorFieldVoronoi} is spatially distributed over the power-Delaunay graph.

As before,  it is possible that under control law \eqref{eq:vectorFieldVoronoi} there exists a time $t
^*$ and $i, j \in I_m$ such that $g_i(t^*) = g_j(t^*)$. Thus, similarly as before, we modify the update equations \eqref{eq:vectorFieldVoronoi} as follows
\begin{equation}\label{eq:vectorFieldVoronoi2}
\begin{split}
\dot{w}_i &= -\frac{\partial \tilde{H}_V}{\partial w_i} -w_i  \text{sat}_{\varepsilon_1,\varepsilon_2} \Bigl ( \|v_{\partial \tilde{H}_i}\| \Bigr) \,\text{sat}_{0,\varepsilon_3} \Bigl (\text{dist}(g_i,\partial V_i)\Bigr) \cdot\prod_{g_j \in M_i(G,\Delta)}\psi\Bigl (\|g_j - g_i \|, \, \,\vartheta_{ij} \Bigr) \doteq u_i^{\text{vor,w}} ,\\
\dot{g}_i &= 
 w_i\frac{\partial \tilde{H}_V}{\partial w_i} \frac{v_{\partial \tilde{H}_i}}{\|\ v_{\partial \tilde{H}_i} \|^2}  \text{sat}_{\varepsilon_1,\varepsilon_2} \Bigl ( \|v_{\partial \tilde{H}_i}\| \Bigr) \,\text{sat}_{0,\varepsilon_3} \Bigl (\text{dist}(g_i,\partial V_i)\Bigr)\cdot \prod_{g_j \in M_i(G,\Delta)}\psi\Bigl (\|g_j - g_i \|, \, \,\vartheta_{ij}\Bigr) \doteq u_i^{\text{vor,g}},
\end{split}
\end{equation}
where $\vartheta_{ij}$ is defined as in Section \ref{sec:powcent}, with $ w_i\frac{\partial \tilde{H}_V}{\partial w_i}v_{\partial \tilde{H}_i}$ replacing $v_{C,g_i}$.

One can prove the following result.
\begin{theorem}\label{thrm:convergenceVoronoi}  Consider the vector field on $\tilde S$ defined by equation \eqref{eq:vectorFieldVoronoi2}. Then generators' positions and weights starting at $t=0$ at $G_W(0) \in \tilde S$, and evolving under \eqref{eq:vectorFieldVoronoi2} remain in $\tilde S$ and converge asymptotically to the set of critical points of the primary objective function $\tilde H_V$ (i.e., to the set of vectors of generators' positions and weights that yield an equitable power diagram).
\end{theorem}
\smallskip
\begin{proof}
Consider $\tilde H_V$ as a Lyapunov function candidate. First, we prove that set $\tilde S$ is positively invariant with respect to \eqref{eq:vectorFieldVoronoi2}. Indeed, by definition of \eqref{eq:vectorFieldVoronoi2}, we have $g_i \neq g_j$ for $i \neq j$ for all $t \geq 0$ (therefore, the power diagram is always well defined). Moreover,  it is straightforward to see that $\dot{\tilde H}_V\leq 0$. Therefore, it holds
$$\lambda_{V_{i}(G_W(t))}^{-1} \le \tilde H_{V}(G_W(t)) \le \tilde H_{V}(G_W(0)), \quad i \in I_m, \, \, t\geq 0.$$ 
Since the measures of the power cells depend continuously on generators'
positions and weights, we conclude that the measures of all power cells
will be bounded away from zero. Finally, since $\dot g_i = 0$ on the
boundary of $A$ for all $i \in I_m$, each generator will remain within the
compact set $A$. Thus, generators' positions and weights will belong to
$\tilde S$ for all $t \geq 0$, that is, $G_W(t)\in \tilde S$ $\forall t\geq 0$.

Second, as shown before, $\tilde H_V:\tilde S \to \reals_{\geq 0}$ is non-increasing along system trajectories, i.e., $\dot{\tilde H}_V(G_W) \leq 0$ in $\tilde S$. 

Third, all trajectories with initial conditions in $\tilde S$ are bounded. Indeed, we have already shown that each generator remains within the compact set $A$ under control law \eqref{eq:vectorFieldVoronoi2}. As far as the weights are concerned, we start by noticing that the time derivative of the sum of the weights is
$$\frac{\partial\sum_{i=1}^m w_i}{\partial t} =  - \sum_{i=1}^m  w_i  \text{sat}_{\varepsilon_1,\varepsilon_2} \Bigl ( \|v_{\partial \tilde{H}_i}\| \Bigr) \,\text{sat}_{0,\varepsilon_3} \Bigl (\text{dist}(g_i,\partial V_i)\Bigr) \prod_{g_j \in M_i(G,\Delta)}\psi\Bigl (\| g_j -g_i  \|, \, \, \vartheta_{ij} \Bigr),$$
since, similarly as in the proof of Theorem \ref{thrm:convergence},
$\sum_{i=1}^m \frac{\partial \tilde H_V}{\partial w_i} = 0$. Moreover, the
magnitude of the difference between any two weights is bounded by a
constant, that is, 
\begin{equation} \label{eq:weightBound}
|w_i -w_j |\leq \alpha \quad \text{for all} \, \, i, \, j \in I_m;
\end{equation}
if, by contradiction, the magnitude of the difference between any two weights could become arbitrarily large, the measure of at least one power cell would vanish, since the positions of the 
generators are confined within $A$. Assume, by the sake of contradiction, that weights' trajectories are unbounded. This means that
$$\forall R>0 \quad \exists t \, \geq 0 \, \, \text{and} \, \, \exists \, j \in I_m  \quad \text{such that} \quad |w_j(t) |\geq R.$$
For simplicity, assume that $w_i(0) = 0$ for all $i\in I_m$ (the extension to arbitrary initial conditions in $\tilde S$ is straightforward). Choose $R = 2m\alpha$ and let $t_2$ be the time instant such that $|w_j(t_2)| = R$, for some $j \in I_m$. Without loss of generality, assume that $w_j(t_2)>0$. Because of constraint \eqref{eq:weightBound}, we  have $\sum_{i=1}^m w_i(t_2) \geq \frac{\alpha}{2}m(3m+1)$. Let $t_1$ be the \emph{last} time before $t_2$ such that $w_j(t) = m\alpha$; because of continuity of trajectories, $t_1$ is well-defined. Then, because of constraint \eqref{eq:weightBound}, we have (i) $\sum_{i=1}^m w_i(t_1) \leq \frac{\alpha}{2}m(3m-1) < \sum_{i=1}^m w_i(t_2) $, \emph{and} (ii) $\frac{\partial \sum_{i=1}^m w_i(t)}{\partial t} \leq 0$ for $t\in [t_1, \,t_2]$ (since $w_j(t)\geq m\alpha$ for all $t \in[t_1, \, t_2]$ and Eq. \eqref{eq:weightBound} imply $\min_{i \in I_m} w_i(t)>0$ for all $t \in [t_1,\,t_2]$); thus, we get a contradiction.

Finally, by Theorem~\ref{thrm:main2}, $\tilde H_V$ is continuously differentiable in $\tilde S$. Hence, by invoking the LaSalle invariance principle (see, for instance, \cite{FB-JC-SM:08}), under the descent flow \eqref{eq:vectorFieldVoronoi2}, generators' positions and weights will converge asymptotically to the \emph{set} of
critical points of $\tilde H_V$, that is not empty by Theorem~\ref{thm:homotopy}. 
\end{proof}
 
\subsection{Obtaining Cells Similar to Regular Polygons}
\label{sec:isoDefect}
In many applications, it is preferable to avoid long and thin subregions. For example, in applications where a mobile agent has to service demands distributed in its own subregion, the maximum travel distance is minimized when the subregion is a circle. Thus, it is of interest to have subregions whose shapes show \emph{circular symmetry}, i.e., that are similar to regular polygons. 

Define, now, the distortion function $L_V:(A\times \reals)^m\setminus \Gamma _{\text{coinc}} \mapsto \reals_{\geq 0}$: $\sum_{i=1}^m \int_{V_i} \|x - g_i \|^2\lambda(x)dx$ 
(where $V_i$ is the $i$-th cell in the corresponding power diagram).
In \cite{Newman:82} it is shown that, when $m$ is large, for the centroidal Voronoi diagram (i.e., centroidal power diagram with equal weights) that minimizes $L_V$, all cells are approximately congruent to a \emph{regular hexagon}, i.e., to a polygon with considerable circular symmetry (see Section \ref{sec:sim} for a more in-depth discussion about circular symmetry). 

Indeed, it is possible to obtain a power diagram that is \emph{close} to a centroidal Voronoi diagram by combining control laws \eqref{eq:vectorField3} and \eqref{eq:vectorFieldVoronoi2}. In particular, we set up the following (spatially-distributed) control law: 
\begin{equation}\label{eq:finalVectorField}
\begin{split}
\dot{w}_i = &u_i^{\text{cent,w}} + u_i^{\text{vor,w}},\\
\dot{g}_i = &u_i^{\text{cent,g}} + u_i^{\text{vor,g}}.
\end{split}
\end{equation}
Combining the results of Theorem \ref{thrm:convergenceCentroidal} and Theorem \ref{thrm:convergenceVoronoi}, we argue that with control law \eqref{eq:finalVectorField} it is possible to obtain equitable power diagrams with cells \emph{similar} to regular polygons, i.e. that show circular symmetry.

\section{Simulations and Discussion} \label{sec:sim}
In this section we verify through simulation the effectiveness of the optimization for the secondary objectives. Due to space constraints, we discuss only control law \eqref{eq:finalVectorField}. We introduce two criteria to judge, respectively, \emph{closeness} to a Voronoi diagram, and circular symmetry of a partition.

\subsection{Closeness to Voronoi Diagrams}
In a Voronoi diagram, the intersection between the bisector of two neighboring generators $g_i$ and $g_j$, and the line segment joining $g_i$ and $g_j$ is the midpoint $g_{ij}^{\text{vor}} \doteq (g_i + g_j)/2$. Then, if we define $g_{ij}^{\text{pow}}$ as the intersection, in a power diagram, between the bisector of two neighboring generators $(g_i, w_i)$ and $(g_j, w_j)$ and the line segment joining their positions $g_i$ and $g_j$, a possible way to measure the \emph{distance} $\eta$ of a power diagram from a Voronoi diagram is the following:
\begin{equation}
\eta \doteq \frac{1}{2N}\sum_{i=1}^m\sum_{j\in N_i} \frac{\|g_{ij}^{\text{pow}} - g_{ij}^{\text {vor}} \|}{0.5\,\gamma_{ij}},
\end{equation}
where $N$ is the number of neighboring relationships and, as before, $\gamma_{ij} = \|g_j - g_i \|$. Clearly, if the power diagram is also a Voronoi diagram (i.e., if all weights are equal), we have $\eta = 0$. We will also refer to $\eta$ as the \emph{Voronoi defect} of a power diagram.

\subsection{Circular Symmetry of a Partition}
A quantitative manifestation of circular symmetry is the well-known \emph{isoperimetric inequality} which states that among all planar objects of a given perimeter, the
circle encloses the largest area. More precisely, given a planar region $V$ with perimeter $p_{V}$ and area $|V|$, then $p_V^2 - 4\pi|V| \geq 0$, and equality holds if and only if $V$ is a circle. Then, we can define the \emph{isoperimetric ratio} as follows: $Q_V = \frac{4\pi |V|}{p_V^2}$;
by the isoperimetric inequality, $Q_V \leq 1$, with equality only for circles. Interestingly, for a regular $n$-gon the isoperimetric ratio $Q_n$ is $Q_n = \frac{\pi}{n \tan\frac{\pi}{n}}$,
which converges to $1$ for $n \to \infty$. Accordingly, given a partition $\mathcal A = \{A_i \}_{i=1}^{m}$, we define, as a measure for the circular symmetry of the partition, the isoperimetric ratio $Q_{\mathcal A} \doteq \frac{1}{m}\sum{Q_{A_i}}$. 

\subsection{Simulation Results}

We simulate ten agents providing service in the unit square $A$. Agents' initial positions are independently and uniformly distributed over $A$, and all weights are initialized to zero. Time is discretized with a step $dt = 0.01$, and each simulation run consists of $800$ iterations (thus, the final time is $T=8$). Define the area error $\epsilon$ as the difference, at $T=8$, between the measure of the region of dominance with maximum measure and the measure of the region of dominance with minimum measure.

First, we consider a measure $\lambda$ \emph{uniform} over $A$, in particular $\lambda  \equiv 1$. Therefore, we have $\lambda_A = 1$ and agents should reach a partition in which each region of dominance has measure equal to $0.1$. For this case, we run $50$ simulations. 

Then, we consider a measure $\lambda$ that follows a gaussian distribution, namely $\lambda(x,y)=e^{-5((x-0.8)^2 + (y-0.8)^2)}$, $(x,y) \in A$, whose peak is at the north-east corner of the unit square. Therefore, we have $\lambda_A \approx 0.336$, and vehicles should reach a partition in which each region of dominance has measure equal to $0.0336$. For this case,  we run $20$ simulations. 

\begin{figure}[thpb]
\centering  
    \mbox{
      \subfigure[Typical equitable partition of $A$ for $\lambda(x,y)=1$.]
      {\scalebox{0.45}{\includegraphics{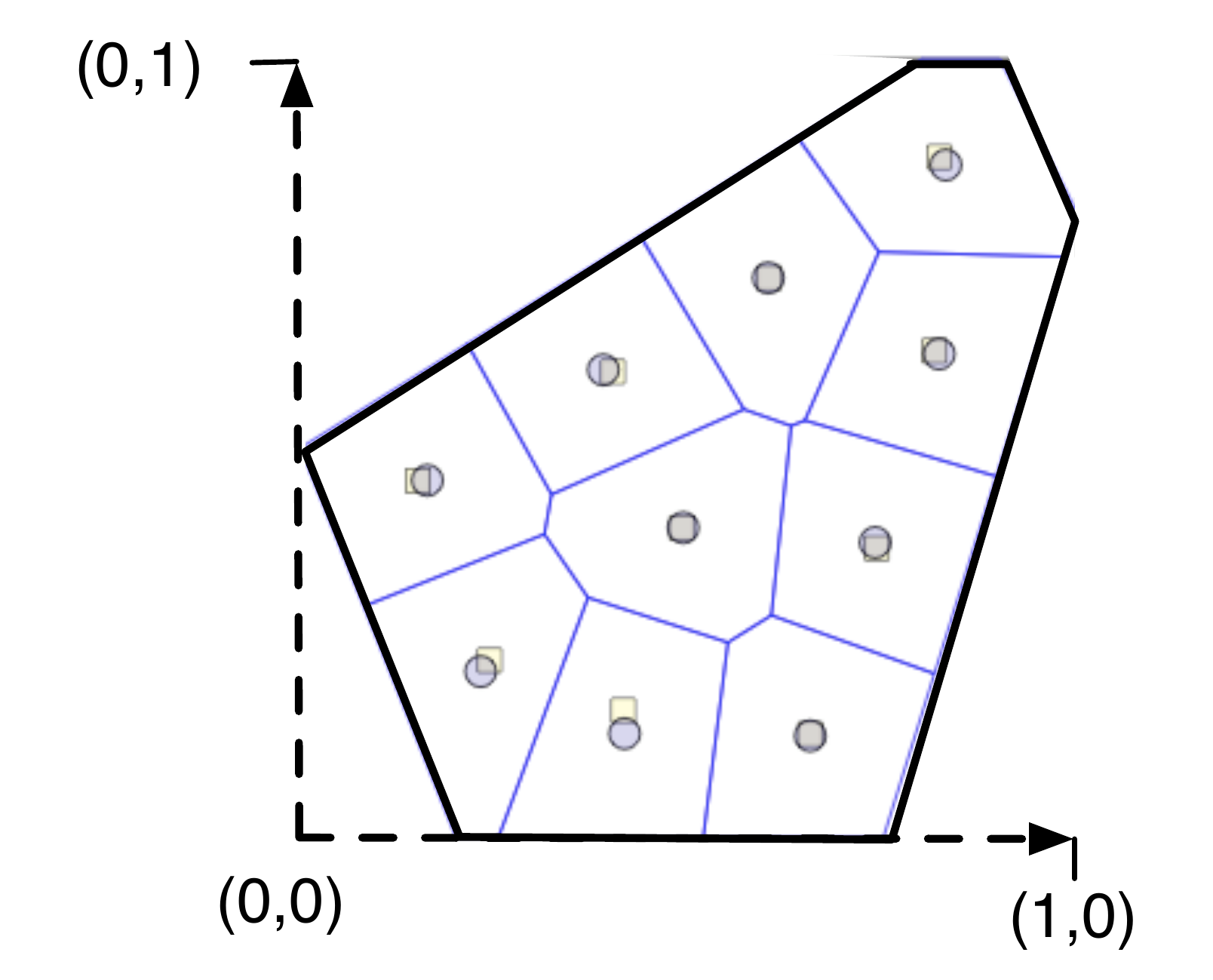}}}
\quad \quad 
      \subfigure[Typical equitable partition of $A$ for $\lambda(x,y)=e^{-5((x-0.8)^2 + (y-0.8)^2)}$.]
      {\scalebox{0.45}{\includegraphics{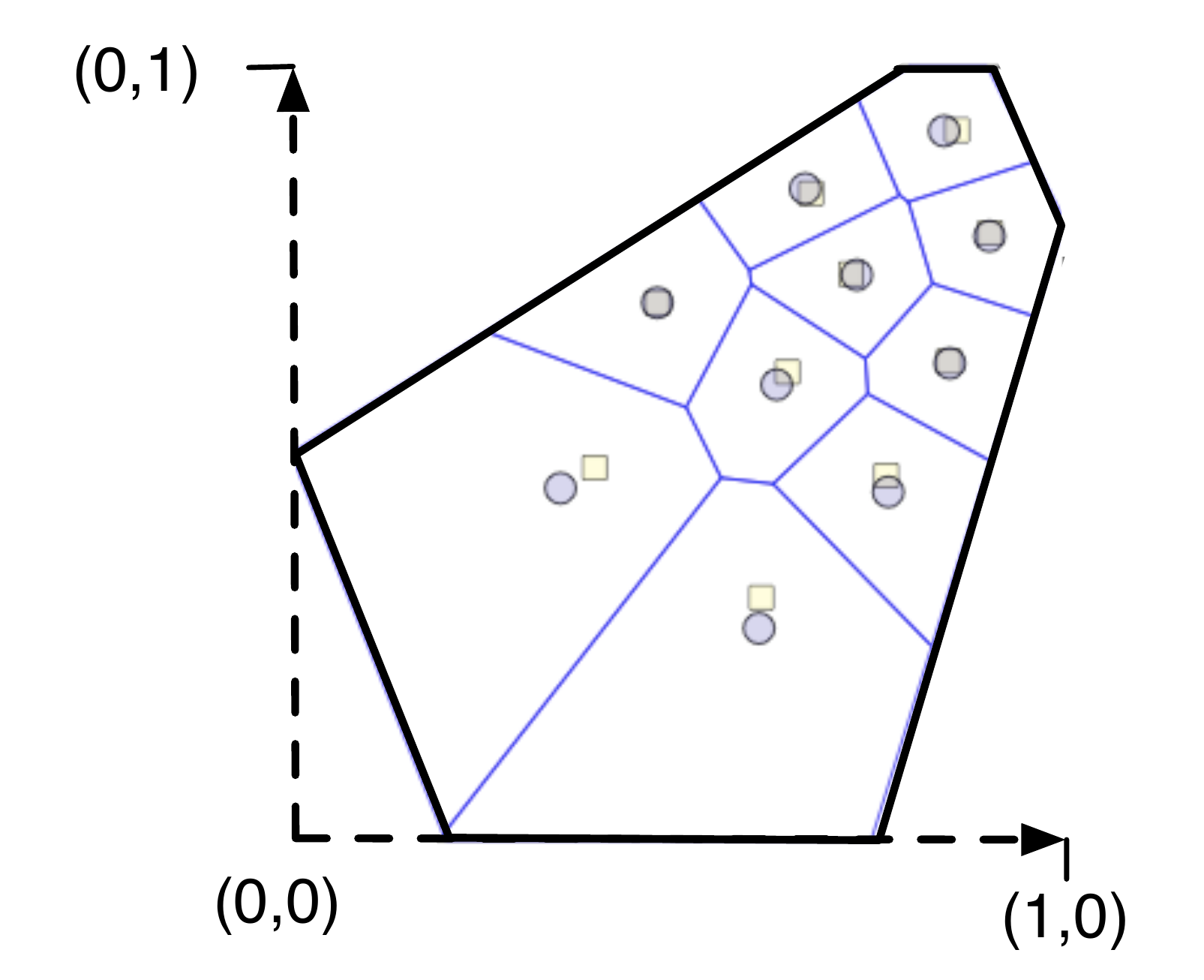}}}

} 
    \caption{Typical equitable partitions achieved by using control law \eqref{eq:finalVectorField}. The yellow squares represent the position of the generators, while the blue circles represent the centroids. Notice how each bisector intersects the line segment joining the two corresponding power neighbors almost at the midpoint; hence both partitions are very close to Voronoi partitions. Compare with Fig. \ref{fig:sliceSweep}.}
    \label{fig:shape}
\end{figure}

Table \ref{tbl:simulation1} summarizes simulation results for the uniform $\lambda$ ($\lambda$=unif) and the gaussian $\lambda$ ($\lambda$=gauss). Expectation and worst case values of area error $\epsilon$, Voronoi defect $\eta$ and isoperimetric ratio 
$Q_{\mathcal V}$ are with respect to $50$ runs for uniform $\lambda$, and $20$ runs for gaussian $\lambda$. Notice that for both measures, after $800$ iterations, (i) the worst case area error is within $16\%$ from the desired measure of dominance regions, (ii) the worst case $\eta$ is very close to $0$, and, finally, (iii) cells have, approximately, the circular symmetry of squares (since $Q_4\approx 0.78$). Therefore, convergence to a convex equitable partition with the desired properties (i.e., closeness to Voronoi diagrams and circular symmetry) seems to be robust. Figure \ref{fig:shape} shows the typical equitable partitions that are achieved with control law  \eqref{eq:finalVectorField} with $10$ agents.

\begin{table}
\caption{Performance of control law \eqref{eq:finalVectorField}.}
\label{tbl:simulation1}
\centering
\begin{tabular}{|c||cccccc|}
        \hline 
$\lambda$ & $\expectation{\epsilon}$   &   $\max \epsilon$   & $\expectation{\eta}$   &   $\max \eta$  & $\expectation{Q_{\mathcal V}}$    &   $\min Q_{\mathcal V}$    \\
        \hline \hline
unif &$3.8\, 10^{-4}$ &$0.016$&$0.01$&$0.03$&$0.73$&$0.66$\\       
 gauss &$3\, 10^{-3}$ &$5.3 \, 10^{-3}$&$0.02$&$0.04$&$0.75$&$0.69$\\    
 \hline
\end{tabular}
\end{table}

\section{Application and Conclusion}
In this last section, we present an application of our algorithms and we draw our conclusions.

\subsection{Application} 
A possible application of our algorithms is in the Dynamic Traveling Repairman Problem (DTRP). In the DTRP, $m$ agents operating in a workspace $A$ must service demands whose time of arrival, location and on-site service are stochastic; the objective is to find a policy to service demands over an infinite horizon that minimizes the expected system time (wait plus service) of the demands. There are many practical settings in which such problem arises. Any distribution
system which receives orders in real time and makes deliveries based on these orders
(e.g., courier services) is a clear candidate. Equitable partitioning policies (with respect to a suitable measure $\lambda$ related to the probability distribution of demand locations) are, indeed, optimal for the DTRP when the arrival rate for service demands is large enough (see \cite{Bertsimas.vanRyzin.Demand:93, Bertsimas.vanRyzin:93b, Xu:95}). Therefore, it is of interest to combine the optimal equitable partitioning policies in \cite{Bertsimas.vanRyzin:93b} with the spatially-distributed algorithms presented in this paper.

\begin{figure}[thpb]
\centering  
    \mbox{
      \subfigure[Agents, virtual generators and regions of dominance.]
      {\label{fig:virtualA}\scalebox{0.4}{\includegraphics{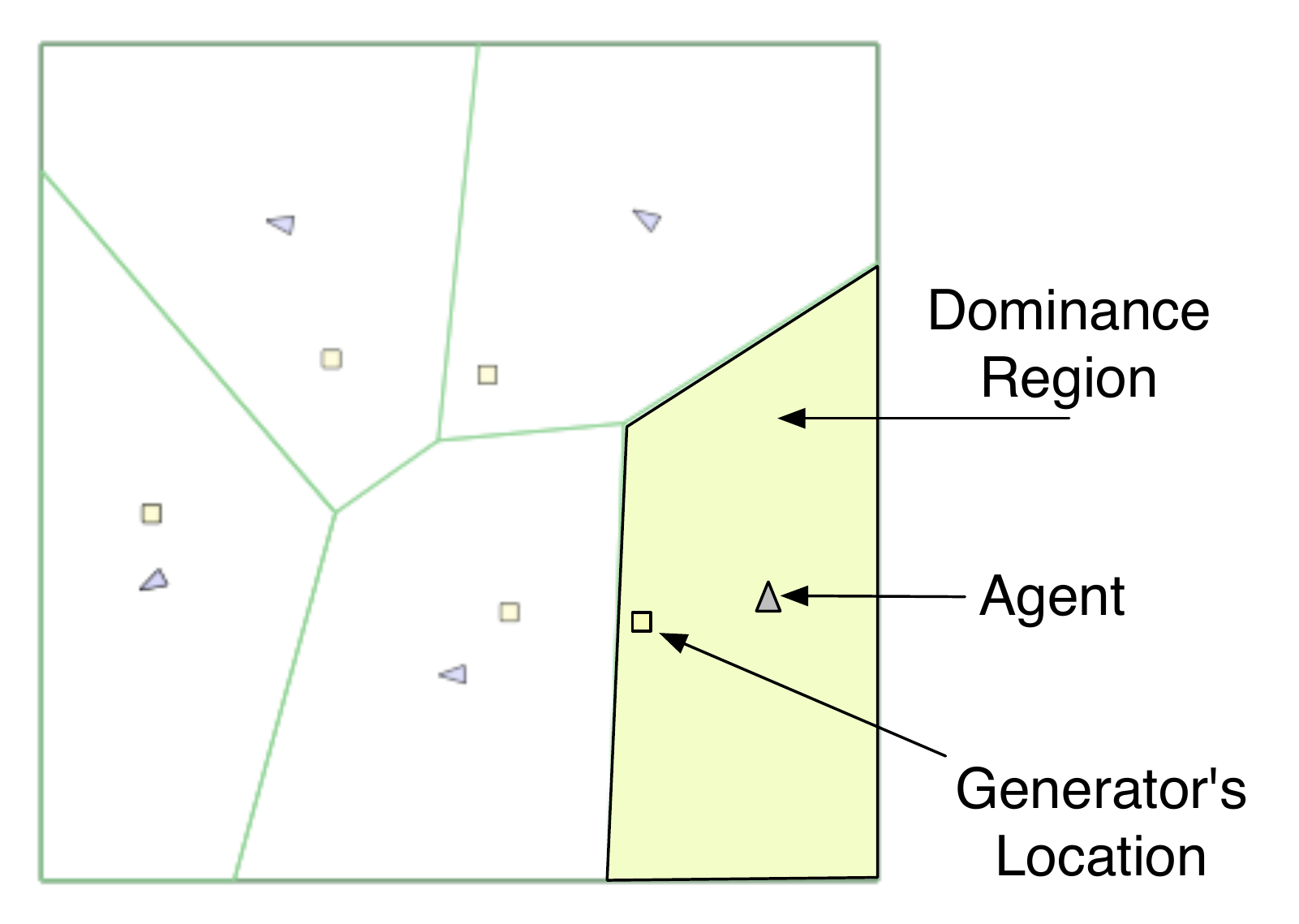}}}
\quad \quad 
      \subfigure[Each agent services outstanding demands inside its own region of dominance.]
      {\label{fig:virtualB}\scalebox{0.4}{\includegraphics{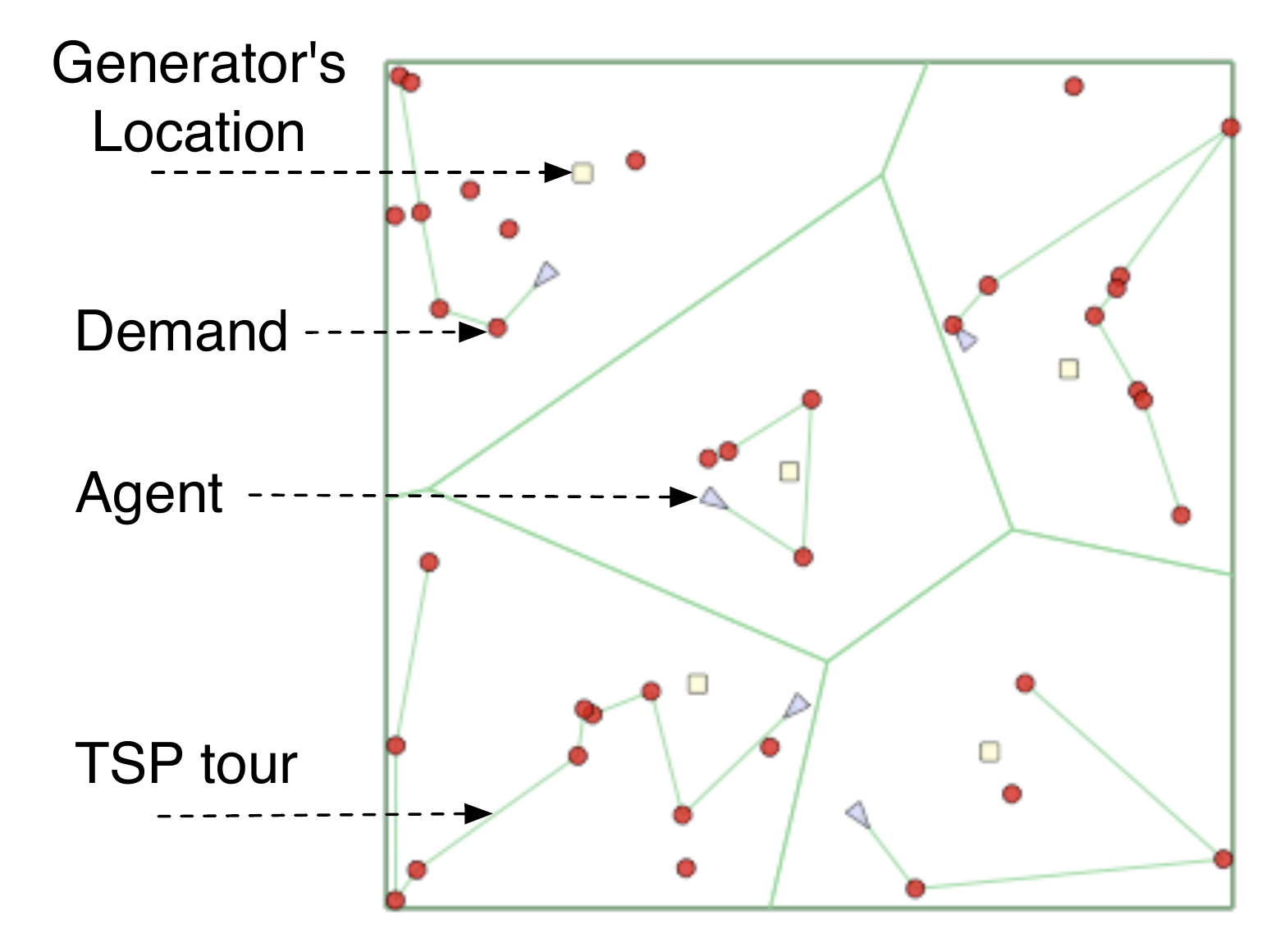}}}

} 
    \caption{Spatially-distributed algorithms for the DTRP.}
    \label{fig:virtualGen}
\end{figure}

The first step is to associate to each agent $i$ a \emph{virtual power
  generator} (virtual generator for short) $(g_i,w_i)$. We define the
region of dominance for agent $i$ as the power cell $ V_i = V_i(G_W)$,
where $G_W = \Bigl((g_1,w_1),\cdots,(g_m,w_m) \Bigr)$ (see Fig. \ref{fig:virtualA}).
We refer
to the partition into regions of dominance induced by the set of virtual
generators $G_W$ as $ \mathcal V(G_W)$. A virtual generator $(g_i, w_i)$ is
simply an artificial variable locally controlled by the $i$-th agent; in
particular, $g_i$ is a virtual point and $w_i$ is its weight. 
Virtual generators allow us to decouple the problem of achieving an
equitable partition into regions of dominance from that of positioning an
agent inside its own region of dominance. 

Then, each agent applies to its virtual generator one of the previous algorithms, while it performs inside its region of dominance the optimal single-agent policy  described in \cite{Bertsimas.vanRyzin.Demand:93} (see Fig. \ref{fig:virtualB}).

%

Notice that, since each agent is required to travel inside its own region of dominance, this scheme is inherently safe against collisions.

\subsection{Conclusion}
We have presented provably correct, spatially-distributed control policies that allow a team of agents to achieve a convex and equitable partition of a convex workspace. We also considered the issue of achieving convex and equitable partitions with special properties (e.g., with hexagon-like cells). Our algorithms could find applications in many
problems, including dynamic vehicle routing, and wireless networks.  This
paper leaves numerous important extensions open for further research.
First, all the algorithms that we proposed are synchronous: we plan to devise algorithms that are amenable to asynchronous implementation. Second, we envision considering the setting of structured
environments (ranging from simple nonconvex polygons to more realistic
ground environments). Finally, to assess the closed-loop robustness and the
feasibility of our algorithms, we plan to implement them on a network of
unmanned aerial vehicles. 

\section*{Acknowledgments}
We gratefully acknowledge Professor A. Bressan's help in deriving the proof
of Theorem \ref{th:Voronoi}.  The research leading to this work was
partially supported by the National Science Foundation through grants
\#0705451 and \#0705453 and by the Office of Naval Research through grant
\# N00014-07-1-0721. Any opinions, findings, and conclusions or
recommendations expressed in this publication are those of the authors and
do not necessarily reflect the views of the supporting organizations.

\ifCLASSOPTIONcaptionsoff
  \newpage
\fi



\bibliographystyle{IEEEtran}
\bibliography{Marco}

\newcommand{\noopsort}[1]{} \newcommand{\printfirst}[2]{#1}
  \newcommand{\singleletter}[1]{#1} \newcommand{\switchargs}[2]{#2#1}
\begin{thebibliography}{10}
\providecommand{\url}[1]{#1}
\csname url@samestyle\endcsname
\providecommand{\newblock}{\relax}
\providecommand{\bibinfo}[2]{#2}
\providecommand{\BIBentrySTDinterwordspacing}{\spaceskip=0pt\relax}
\providecommand{\BIBentryALTinterwordstretchfactor}{4}
\providecommand{\BIBentryALTinterwordspacing}{\spaceskip=\fontdimen2\font plus
\BIBentryALTinterwordstretchfactor\fontdimen3\font minus
  \fontdimen4\font\relax}
\providecommand{\BIBforeignlanguage}[2]{{%
\expandafter\ifx\csname l@#1\endcsname\relax
\typeout{** WARNING: IEEEtran.bst: No hyphenation pattern has been}%
\typeout{** loaded for the language `#1'. Using the pattern for}%
\typeout{** the default language instead.}%
\else
\language=\csname l@#1\endcsname
\fi
#2}}
\providecommand{\BIBdecl}{\relax}
\BIBdecl

\bibitem{Bertsimas.vanRyzin.Demand:93}
D.~J. Bertsimas and G.~J. {van~Ryzin}, ``Stochastic and dynamic vehicle routing
  in the {E}uclidean plane with multiple capacitated vehicles,'' \emph{Advances
  in Applied Probability}, vol.~25, no.~4, pp. 947--978, 1993.

\bibitem{Baron.Berman.ea:07}
O.~Baron, O.~Berman, D.~Krass, and Q.~Wang, ``The equitable location problem on
  the plane,'' \emph{European Journal of Operational Research}, vol. 183,
  no.~2, pp. 578--590, 2007.

\bibitem{Liu.Liu.ea:03}
B.~Liu, Z.~Liu, and D.~Towsley, ``On the capacity of hybrid wireless
  networks,'' in \emph{IEEE INFOCOM 2003}, San Francisco, CA, Apr. 2003, pp.
  1543--1552.

\bibitem{Carlsson.Ge.ea:07}
J.~Carlsson, D.~Ge, A.~Subramaniam, A.~Wu, and Y.~Ye, ``Solving min-max
  multi-depot vehicle routing problem,'' \emph{Report}, 2007.

\bibitem{Lloyd:82}
S.~P. Lloyd, ``Ò{L}east-squares quantization in {PCM},'' \emph{IEEE Trans.
  Information Theory}, vol.~28, no.~2, pp. 129--137, 1982.

\bibitem{FB-JC-SM:08}
F.~Bullo, J.~Cort\'es, and S.~Mart{\'\i}nez, \emph{Distributed Control of
  Robotic Networks}, ser. Applied Mathematics Series.\hskip 1em plus 0.5em
  minus 0.4em\relax Princeton University Press, Sep. 2008, manuscript under
  contract. Electronically available at http://coordinationbook.info.

\bibitem{Kwok.Martinez:07}
A.~Kwok and S.~Martinez, ``Energy-balancing cooperative strategies for sensor
  deployment,'' in \emph{Proc.\ IEEE Conf.\ on Decision and Control}, New
  Orleans, LA, Dec. 2007, pp. 6136--6141.

\bibitem{Carlsson.Armbruster.ea:08}
J.~Carlsson, B.~Armbruster, and Y.~Ye, ``Finding equitable convex partitions of
  points in a polygon efficiently,'' \emph{To appear in The ACM Transactions on
  Algorithms}, 2008.

\bibitem{Diestel:00}
R.~Diestel, \emph{Graph Theory}, 2nd~ed., ser. Graduate Texts in
  Mathematics.\hskip 1em plus 0.5em minus 0.4em\relax New York: Springer
  Verlag, 2000, vol. 173.

\bibitem{Chavel:84}
I.~Chavel, \emph{Eigenvalues in Riemannian Geometry}.\hskip 1em plus 0.5em
  minus 0.4em\relax New York, NY: Academic Press, 1984.

\bibitem{Hatcher:01}
\BIBentryALTinterwordspacing
A.~Hatcher, \emph{Algebraic Topology}.\hskip 1em plus 0.5em minus 0.4em\relax
  Cambridge, U.K.: Cambridge University Press, 2002. [Online]. Available:
  \url{http://www.math.cornell.edu/$\sim$hatcher/AT/ATpage.html}
\BIBentrySTDinterwordspacing

\bibitem{Okabe:00}
A.~Okabe, B.~Boots, K.~Sugihara, and S.~N. Chiu, \emph{Spatial Tessellations:
  Concepts and Applications of Voronoi Diagrams}.\hskip 1em plus 0.5em minus
  0.4em\relax New York, NY: John Wiley \& Sons, 2000.

\bibitem{Imai.Iri.ea:85}
H.~Imai, M.~Iri, and K.~Murota, ``Voronoi diagram in the {L}aguerre geometry
  and its applications,'' \emph{SIAM Journal on Computing}, vol.~14, no.~1, pp.
  93--105, 1985.

\bibitem{Aurenhammer:87}
F.~Aurenhammer, ``Power diagrams: properties, algorithms and applications,''
  \emph{SIAM Journal on Computing}, vol.~16, no.~1, pp. 78--96, 1987.

\bibitem{JC-SM-FB:03p}
J.~Cort{\'e}s, S.~Mart{\'\i}nez, and F.~Bullo, ``Spatially-distributed coverage
  optimization and control with limited-range interactions,'' \emph{{ESAIM.}
  Control, Optimisation \& Calculus of Variations}, vol.~11, pp. 691--719,
  2005.

\bibitem{Pavone.Frazzoli.ea:07}
M.~Pavone, E.~Frazzoli, and F.~Bullo, ``Decentralized algorithms for stochastic
  and dynamic vehicle routing with general demand distribution,'' in
  \emph{Proc.\ IEEE Conf.\ on Decision and Control}, New Orleans, LA, Dec.
  2007, pp. 4869--4874.

\bibitem{Pavone.Frazzoli.ea:CDC08}
------, ``Distributed policies for equitable partitioning: Theory and
  applications,'' in \emph{Proc.\ IEEE Conf.\ on Decision and Control}, Cancun,
  Mexico, Dec. 2008.

\bibitem{Newman:82}
D.~Newman, ``The hexagon theorem,'' \emph{IEEE Transactions on Information
  Theory}, vol.~28, no.~2, pp. 137--139, Mar 1982.

\bibitem{Bertsimas.vanRyzin:93b}
D.~J. Bertsimas and G.~J. {van~Ryzin}, ``Stochastic and dynamic vehicle routing
  with general interarrival and service time distributions,'' \emph{Advances in
  Applied Probability}, vol.~25, pp. 947--978, 1993.

\bibitem{Xu:95}
H.~Xu, ``Optimal policies for stochastic and dynamic vehicle routing
  problems.'' Dept. of {C}ivil and {E}nvironmental {E}ngineering, Massachusetts
  Institute of Technology, Cambridge, MA., 1995.

\bibitem{Bressan:08}
A.~Bressan, \emph{Personal Communication}, 2008.

\end{thebibliography}

\section*{Appendix}

\begin{proof}[Proof of Theorem \ref{th:Voronoi}]
The proof mainly relies on \cite{Bressan:08}. Let $v$ be the unit vector considered in the definition of the Unimodal Property. Then, there exist unique values
$s_0<s_1<\cdots<s_m $
such that 
$s_0 = \inf\{s;  A^s \neq \emptyset \}$, 
$s_m = \sup\{s;  A^s \neq \emptyset \}$, and
\begin{equation}\label{eq:slice} 
\lambda_{  \{ x\in  A; \, v\cdot x\leq s_k\} }=\frac{k}{m} \lambda_{A}, \quad k=1,\ldots,m-1.
\end{equation} 
Consider the intervals
$I_i \doteq [s_{i-1},s_i]$, $i \in I_m. $
We claim that one can choose points
$g_i=t_i v\in \reals^d$, $ i\in I_m  $
such that $t_i \in I_i$ and the corresponding Voronoi diagram is 
\begin{equation}\label{eq:vorSlice}
\begin{split}
 A_i &= \{x \in  A; \quad \|x - g_i\| = \min_k\|x-g_k\| \}\\
&=\{x \in  A; \quad v\cdot x\in [s_{i-1},s_i] \}.
\end{split}
\end{equation}
Together, Eq. \eqref{eq:slice} and Eq. \eqref{eq:vorSlice} yield the desired result.

Since, by assumption, $ A$ enjoys the Unimodal Property, there exists an index $\kappa \in \{ 1,\ldots,m\}$ such that the length of the intervals $I_i = [s_{i-1}, s_{i}]$ decreases as $i$ ranges from $1$ to $\kappa$, then increases as $i$ ranges from $\kappa$ to $m$. Let $I_{\kappa} = [s_{\kappa-1},s_{\kappa}]$ be the smallest of these intervals, and define 
$$t_{\kappa}\doteq \frac{s_{\kappa-1}+s_{\kappa}}{2} \in I_{\kappa}. $$
By induction, for $i$ increasing from $\kappa$ to $m-1$, define $t_{i+1}$ be the symmetric to $t_i$ with respect to $s_i$, so that
$$ t_{i+1} = 2s_i-t_i \quad i=\kappa,\kappa+1,\ldots,m-1.$$
Since the length of $I_{i+1}$ is larger than the length of $I_i$, we have 
\begin{equation}\label{eq:ind1}
t_i \in I_i \Rightarrow t_{i+1} \in I_{i+1}.
\end{equation} 
Similarly, for $i$ decreasing from $\kappa$ to $1$, we define
$$t_{i-1}=2s_{i-1}-t_i, \quad i=\kappa,\kappa-1,\ldots,2 .$$ 
Since the interval $I_{i-1}$ is now larger than the interval $I_{i}$, we have 
\begin{equation}\label{eq:ind2}
t_i \in I_i \Rightarrow t_{i-1}\in I_{i-1}.
\end{equation}
By Eqs. \eqref{eq:ind1}-\eqref{eq:ind2} imply $t_i \in I_i$ for all $i=1,\ldots,m$. Hence the second equality in Eq. \eqref{eq:vorSlice}  holds. 
\end{proof}
We now specialize the theorem to the case when $ A$ is convex.
\begin{corollary}\label{cor:Vor}
Let $ A \subset \reals^d$ be a compact, convex set, and $\lambda$ be constant on $A$. Then for every $m \geq1$ there exist points $g_1,\cdots,g_m$ all in the interior of $ A$, such that the corresponding Voronoi diagram is equitable.
\end{corollary}
\begin{proof} Notice that every compact convex set enjoys the Unimodal Property, with an arbitrary choice
of the unit vector $v$. By compactness, there
exist points $a,\, b \in A$ such that $\|b-a\| = \max_{z,z'\in A} \|z - z' \| .$
By a translation of coordinates, we can assume $a$ = 0. Choose $v \doteq b/\|b\|$. Then the previous
construction yields an equitable Voronoi diagram generated by $m$ points $g_i = t_i v$ all in the interior of $ A$.
\end{proof}

\begin{proof}[Proof of Lemma \ref{lemma:Wcont}]
By Theorem \ref{thrm:convergence} and by its very definition $W^*(G)$ is the zero of the vector field $-\frac{\partial H_V}{\partial w_i}(W(t))$. Now let us denote with 
$$K(W,G)\dot=-\frac{\partial H_V}{\partial w_i}(W),$$
the corresponding continuous function, viewed as a function of two independent set of variables, namely the weights $(w_1,\dots,w_n)=W$ and the non-degenerate vector of generators' locations $G$.
In order to prove that the assignment $G\mapsto W^*(G)$ is continuous, notice that by Theorem \ref{thrm:main} the function $K(W,G)$ is identically zero when restricted to the graph of $W^*$, namely $K(W^*(G),G)=0$. The function $W^*$ is continuous iff it is continuous in each of its argument. Fix, first, a generator $g_i \notin \partial\Gamma$ and consider for any $v\in \mathbb{R}^2$, the variation $(g_1,\dots, g_{i-1},g_i+hv,g_{i+1},\dots, g_m)$. Since $g_i \notin \partial\Gamma$, there always exists
 an $\epsilon>0$, depending on $g_i$ and $v$, such that for any $h$ with $0\leq h <\epsilon$, $(g_1,\dots, g_{i-1},g_i+hv,g_{i+1},\dots, g_m)\in \Gamma$. 
Now $K(W^*(g_1,\dots, g_{i-1},g_i+hv,g_{i+1},\dots, g_m),(g_1,\dots, g_{i-1},g_i+hv,g_{i+1},\dots, g_m))=0$ for any $0\leq h<\epsilon$ by definition. Therefore, taking the limit for $h\rightarrow 0^+$, we still get zero. On the other hand, since $K$ is continuous, we can take the limit inside $K$ and we get
$$K(\lim_{h\rightarrow 0^+}W^*(g_1,\dots, g_{i-1},g_i+hv,g_{i+1},\dots), (g_1,\dots, g_{i-1},g_i,g_{i+1},\dots))=0.$$
Therefore, we have that $\lim_{h\rightarrow 0^+}W^*(g_1,\dots, g_{i-1},g_i+hv,g_{i+1},\dots, g_m)$ is equal to $W^*(g_1,\dots, g_{i-1},g_i,g_{i+1},\dots, g_m)$, by the uniqueness in $\Omega$ of the value of $W^*$ for which, given $G$, the function $K$ vanishes. 
\end{proof}

\end{document}